\documentclass{article}

 \usepackage[preprint]{neurips_2026}

\usepackage[utf8]{inputenc} % allow utf-8 input
\usepackage[T1]{fontenc}    % use 8-bit T1 fonts
\usepackage{hyperref}       % hyperlinks
\usepackage{url}            % simple URL typesetting
\usepackage{booktabs}       % professional-quality tables
\usepackage{amsfonts}       % blackboard math symbols
\usepackage{nicefrac}       % compact symbols for 1/2, etc.
\usepackage{microtype}      % microtypography
\usepackage{xcolor}         % colors

\usepackage{graphicx}
\usepackage{amsmath}
\usepackage{multirow}
\usepackage{arydshln}
\usepackage{subcaption}
\usepackage{colortbl}
\usepackage{makecell}
\usepackage{pifont} % for check and x marks
\usepackage{wrapfig}

\newcommand{\proposal}{\textit{ViNUSS}}

\usepackage[normalem]{ulem}
\newcommand{\cmark}{\ding{51}}
\newcommand{\xmark}{\textcolor{gray}{--}}

\usepackage{enumitem}

\title{Subject-Relative Micro-Motion and Sleep Dynamics for Near-Infrared Video Sleep Staging}

\author{Kunmin Jang\thanks{These authors contributed equally.}\\ 
Seoul National University \\ 
\texttt{jangk10@snu.ac.kr} \\ 
\And You Rim Choi\footnotemark[1] \\ 
Seoul National University \\ 
\texttt{yrchoi@snu.ac.kr} \\ 
\And Hun Heo \\ 
Seoul National University \\ 
\texttt{greathunh@snu.ac.kr} \\ 
\And Heonjun Lee \\ 
Seoul National University \\ 
\texttt{johnlee0321@snu.ac.kr} \\ 
\And Suahn Bae \\ 
Seoul National University \\ 
\texttt{snubbb@snu.ac.kr} \\
\And Dongik Park \\ 
Seoul National University \\ 
\texttt{breadsora@snu.ac.kr} \\  
\And Hyun-Woo Shin\thanks{Corresponding authors.} \\ 
Seoul National University Hospital \\ 
\texttt{charlie@snu.ac.kr} \\ 
\And Hyung-Sin Kim\footnotemark[2] \\ 
Seoul National University \\ 
\texttt{hyungkim@snu.ac.kr} \\ }

\begin{document}

\maketitle

\begin{abstract}
Near-infrared (NIR) video is a promising modality for contactless sleep monitoring, but recent video-based sleep staging methods often use it as a route to reconstructed respiratory/cardiac proxies or cross-modal physiological representations. 
We study video-only sleep staging under labels defined by polysomnography (PSG), where the model infers sleep stages from NIR video alone without explicit physiological proxy reconstruction or auxiliary physiological signal supervision. 
This tests whether NIR video itself can provide informative sleep-stage evidence, rather than only serving as an input for recovering physiological proxies. 
We propose \proposal{} (Video-Native Unmediated Sleep Staging), a framework that combines subject-relative micro-motion learning with full-night sleep dynamics modeling. 
Spatially anchored pre-spatial micro-motion encoding preserves localized temporal variation together with its spatial context. 
Within-subject stage contrast learns stage cues with respect to each subject's night-specific baseline. 
Two-scale sleep dynamics modeling captures within-epoch motion evolution and organizes epoch-level evidence into a coherent full-night sleep-stage trajectory. 
On 475 overnight NIR recordings ($\sim$3,250 hours), \proposal{} achieves 0.80 accuracy and 0.78 macro-F1 for four-class sleep staging. 
Interpretability analysis suggests attention to thoraco-abdominal periodic motion and gross body movements associated with arousals and position changes. 
These results support NIR video as an independently informative and complementary modality for PSG-defined sleep-stage estimation.
\end{abstract}    
\section{Introduction}
\label{sec:intro}

Sleep staging is a central procedure in sleep medicine, assigning consecutive 30-second epochs of an overnight recording to Wake, Light (N1/N2), Deep (N3), or REM~\cite{patel2024physiology,AASM}. 
The clinical reference is polysomnography (PSG), where trained technicians score multimodal physiological signals such as EEG, EOG, and EMG~\cite{rundo2019polysomnography}. 
Although PSG provides the standard definition of sleep stages, it requires contact sensors, overnight instrumentation, and expert scoring, limiting scalability for longitudinal and home-based monitoring~\cite{younes2016staging,miettinen2018home}. 
These limitations have motivated contactless approaches that estimate sleep stages under PSG-defined labels while reducing sensing burden.

Near-infrared (NIR) video is a promising modality for contactless sleep monitoring because it operates in darkness and captures spatially localized body motion, including respiration-related motion, limb movement, position changes, and brief arousals. 
Recent video-based sleep staging methods often use video as a route to another representation, such as reconstructed respiratory/cardiac signals, optical-flow-derived motion features, or EEG-guided cross-modal features~\cite{nochino2019sleep,van2023contactless,carter2024sleepvst,wang2024video,yu2025ir_based,han2024iotv2e}.
Proxy-first approaches are valuable and interpretable, but blankets, position changes, and face anonymization can make respiratory or rPPG extraction unreliable, while one-dimensional proxies may discard regional motion patterns and transient movements. 
Cross-modal physiological approaches are complementary, but they answer a different question: how video can align with auxiliary physiological representations.

%EEG-guided approaches are complementary but rely on auxiliary physiological representations beyond PSG labels.

In this work, we study video-only sleep staging under labels defined by PSG. 
The model infers sleep stages from NIR video alone, without explicit physiological proxy reconstruction or auxiliary physiological representation learning.
This formulation does not replace PSG as the clinical reference.
Instead, it tests whether NIR video itself can provide informative sleep-stage evidence, rather than only serving as an input for recovering physiological proxies or aligning to representations derived from other physiological modalities.

This setting differs from generic video understanding. 
Standard video models are typically designed for actions, objects, and scene changes, where discriminative cues often appear as large motion or semantic interaction~\cite{kondratyuk2021movinets,x3d,swin,bertasius2021space,tong2022videomae}. 
Action-oriented temporal modules further enhance motion representation through temporal shifts, motion excitation, or temporal differences~\cite{lin2019tsm,li2020tea,wang2021tdn}. 
In NIR sleep video, however, the scene is mostly static, useful cues may appear as weak local micro-motion, and subject-specific factors such as body shape, bedding, camera geometry, and baseline respiration-related motion can dominate the representation. 
The interpretation of motion also depends on its location, since weak thoraco-abdominal periodicity, fast limb or body movements, and broad position changes provide different sleep-stage evidence. 
Effective video-only sleep staging therefore requires preserving localized temporal variation with its spatial context, learning stage evidence relative to each subject's night-specific baseline, and modeling sleep dynamics across the full night.

We propose \proposal{}, a framework for subject-relative micro-motion learning and full-night sleep dynamics modeling. 
First, spatially anchored pre-spatial micro-motion encoding densely scans the full 30-second epoch with short temporal filters before spatial aggregation, retaining the spatial context of each temporal change. 
This design targets weak, spatially localized traces of sleep-relevant motion over the clinically defined epoch before they are diluted by static appearance cues. 
Second, within-subject stage contrast learns stage evidence relative to the subject's own visual and motion baseline rather than as an absolute appearance template. 
Third, two-scale sleep dynamics modeling captures motion evolution within each 30-second epoch and organizes epoch-level evidence into a coherent full-night sleep-stage trajectory.

Evaluation is conducted on the Korea Video Sleep Study (KVSS)~\cite{choi2026non}, which contains 475 overnight NIR recordings, approximately 3,250 hours of video, synchronized with PSG labels. 
The recordings include far-field viewpoints, bedding occlusion, weak motion, and face anonymization, making them challenging for both proxy extraction and raw-video modeling. 
\proposal{} achieves 0.80 accuracy and 0.78 macro-F1 for four-class staging. 
Ablations show complementary benefits from micro-motion representation, within-subject stage contrast, and full-night sleep dynamics. 
Sleep Video EigenCAM (SVECAM) further suggests that \proposal{} uses physiologically meaningful motion, including thoraco-abdominal periodicity and larger movements associated with arousals and position changes.

Our contributions are summarized as follows:
\begin{itemize}[leftmargin=*, noitemsep, topsep=0pt]
    \item \textbf{Video-only formulation under PSG-defined labels.}
    We study NIR video sleep staging where PSG provides the clinical labels but the model uses NIR video alone, without explicit physiological proxy reconstruction or EEG-guided representation learning.

    \item \textbf{Subject-relative micro-motion learning.}
    We introduce pre-spatial micro-motion encoding and within-subject stage contrast to preserve sleep-relevant temporal traces over the 30-second epoch while learning stage evidence relative to each subject's night-specific baseline.

    \item \textbf{Two-scale sleep dynamics.}
    We model within-epoch motion evolution and full-night temporal context, organizing video-derived epoch evidence into coherent overnight sleep-stage trajectories.

    \item \textbf{Large-scale evaluation and visual evidence analysis.}
    We evaluate \proposal{} on 475 overnight NIR recordings and analyze physiologically meaningful visual evidence.
\end{itemize}
\section{Related Work}
\label{sec:related_work}

%%%%%%%%%%%%%%%%%%%%%%%%%%%%%%%%%%%%%%%%%%%%%%%%%%%%%%
%%%%%%%%%%%%%%%%%%%%%%%%%%%%%%%%%%%%%%%%%%%%%%%%%%%%%%
\subsection{Sleep Staging and Contactless Monitoring}

The clinical reference for sleep staging is polysomnography (PSG), where trained technicians score multi-channel physiological signals such as EEG, EOG, and EMG according to the American Academy of Sleep Medicine (AASM) guidelines~\cite{AASM}. 
Although PSG provides high-fidelity labels, it requires contact sensors, overnight instrumentation, and expert scoring, which limits scalability for longitudinal monitoring~\cite{younes2016staging}.
Wearable devices based on actigraphy, PPG, or simplified EEG reduce this burden~\cite{fonseca2023contactppg,casciola2021deep}, but still require physical contact and can be affected by comfort, adherence, and motion artifacts.
Fully contactless sensing, including audio~\cite{hong2022end}, under-mattress pressure~\cite{vyas2021sleep}, and radio-frequency methods~\cite{hong2018noncontact}, avoids body-worn sensors but often captures only limited aspects of sleep behavior.
Among contactless modalities, near-infrared (NIR) video is particularly relevant because it captures spatially localized body motion in darkness.

%%%%%%%%%%%%%%%%%%%%%%%%%%%%%%%%%%%%%%%%%%%%%%%%%%%%%%
%%%%%%%%%%%%%%%%%%%%%%%%%%%%%%%%%%%%%%%%%%%%%%%%%%%%%%
\subsection{Video-Based and Cross-Modal Sleep Staging}

Video-based sleep staging has been studied through handcrafted motion features and proxy-first pipelines. 
Early camera-based methods used frame-difference features with SVMs~\cite{nochino2019sleep}, while recent methods estimate physiological proxies before classification. 
For example, van Meulen \textit{et al.}~\cite{van2023contactless} used camera-based PPG and heart-rate variability, Video-PSG~\cite{wang2024video} extracted respiratory and cardiac signals, and SleepVST~\cite{carter2024sleepvst} modeled respiratory/cardiac waveforms and optical-flow features with pretrained transformers. 
These proxy-first methods are valuable because they provide physiologically interpretable inputs. 
However, explicit proxy reconstruction can be unreliable in realistic NIR recordings: bedding and position changes can corrupt camera-based respiratory motion~\cite{wang2022algorithmic}, while far-field views, occlusion, and face anonymization can weaken rPPG-based cardiac estimation~\cite{McDuff2023camera,bhutani2025rppgprivacy,Gupta2023privacy}. 
Moreover, proxy compression can discard regional motion patterns and transient movements that are not captured by predefined physiological signals.

Recent work has also connected infrared video with EEG-derived knowledge through EEG-teacher distillation~\cite{yu2025ir_based} or cross-modal retrieval between infrared videos and EEG representations~\cite{han2024iotv2e}.
These studies show that IR video contains sleep-relevant visual evidence when guided by physiological supervision.
Our work is complementary: PSG defines the labels, but our model does not use EEG-teacher features, EEG retrieval databases, or explicit respiratory/cardiac proxy reconstruction.
We instead study NIR video as an independently informative modality through subject-relative micro-motion learning and full-night video-conditioned dynamics.

%%%%%%%%%%%%%%%%%%%%%%%%%%%%%%%%%%%%%%%%%%%%%%%%%%%%%%
%%%%%%%%%%%%%%%%%%%%%%%%%%%%%%%%%%%%%%%%%%%%%%%%%%%%%%
\subsection{Video Understanding and Full-Night Temporal Modeling}

Modern video understanding has developed strong spatiotemporal representations for action recognition and semantic video learning, including efficient 3D CNNs, spatiotemporal transformers, and self-supervised video encoders~\cite{kondratyuk2021movinets,x3d,swin,bertasius2021space,tong2022videomae}. 
Temporal modules further improve action recognition by exchanging information across frames, exciting motion-sensitive channels, or modeling temporal differences~\cite{lin2019tsm,li2020tea,wang2021tdn}. 
These methods motivate motion-aware video processing, but they primarily target salient action cues, whereas NIR sleep video requires preserving dense, low-amplitude temporal traces over a clinically defined 30-second epoch before spatial abstraction.

Sleep staging also requires temporal context beyond isolated epochs, and prior PSG/EEG-based neural models commonly exploit neighboring-epoch structure~\cite{supratak2017deepsleepnet,phan2021xsleepnet,phan2022sleeptransformer}. 
For NIR video, the evidence is indirect and subject-dependent, making longer temporal context important for interpreting local motion. 
\proposal{} therefore summarizes motion within each 30-second epoch and then models the resulting representations across the subject's full-night trajectory using state-space sequence modeling~\cite{gu2023mamba}. 
This couples full-night temporal context with pre-spatial micro-motion encoding and within-subject stage contrast.
\section{Method}
\label{sec:method}

Each overnight recording is divided into 30-second \textit{epochs}, following clinical sleep-staging practice.
Given a full-night NIR video sequence
$\mathcal{V}=\{X_1,\ldots,X_N\}$ with $N$ epochs, our goal is to predict a stage sequence
$\hat{\mathcal{Y}}=\{\hat{y}_1,\ldots,\hat{y}_N\}$ for ground-truth labels
$\mathcal{Y}=\{y_1,\ldots,y_N\}$, where
$y_t \in \{\text{Wake}, \text{Light}, \text{Deep}, \text{REM}\}$.
The labels are defined by PSG scoring, but the model input is NIR video only.
We do not reconstruct respiratory or cardiac waveforms, nor use auxiliary physiological signals such as EEG as model inputs or for cross-modal supervision.

%%%%%%%%%%%%%%%%%%%%%%%%%%%%%%%%%%%%%%%%%%%%%%%%%%%%%%
%%%%%%%%%%%%%%%%%%%%%%%%%%%%%%%%%%%%%%%%%%%%%%%%%%%%%%
\subsection{Design Motivation and \proposal{} Overview}
\label{sec:motivation}

\begin{figure}[t]
    \centering
    \includegraphics[width=\linewidth]{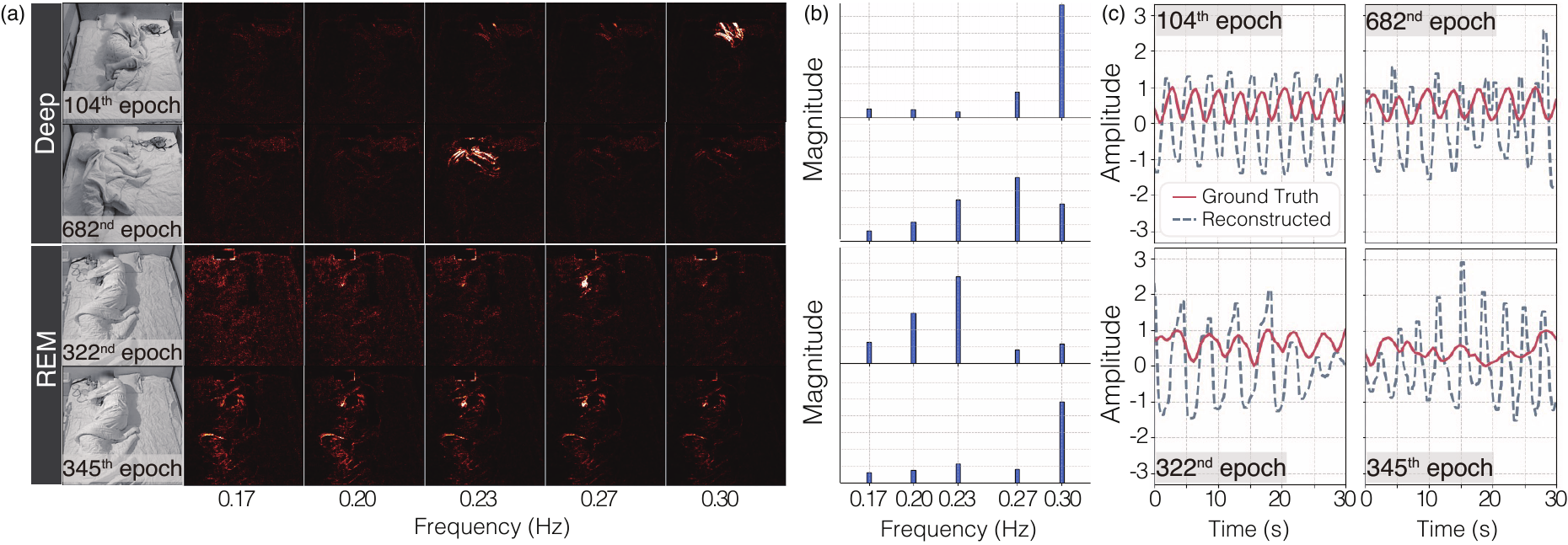}
    \caption{
    Frequency analysis and respiratory reconstruction from NIR sleep video.
    (a) \textit{Video frequency maps} from pixel-wise FFT over 0.17--0.30~Hz; brighter regions indicate stronger periodic motion.
    (b) \textit{Contact thoracic spectra} from chest-belt signals.
    (c) \textit{Video-reconstructed respiratory waveforms}.
    }
    % \vspace{-1.5ex}
    \label{fig:preliminary}
\end{figure}

Figure~\ref{fig:preliminary} illustrates why explicit proxy reconstruction is informative but incomplete in realistic NIR sleep videos.
We reconstruct respiratory waveforms using video motion magnification and camera-based respiratory extraction methods~\cite{wadhwa2013phase,wang2022algorithmic}. 
Figure~\ref{fig:preliminary}(a) shows localized periodic responses around the thoraco-abdominal region, with different dominant frequencies across epochs, indicating respiration-related video evidence. 
However, panel (b) shows that video-dominant frequencies can deviate from contact thoracic spectra under position changes, occlusion, weak motion, or non-respiratory body movement, and panel (c) shows attenuation or frequency mismatch in waveforms.
For example, the Deep \(682^{\mathrm{nd}}\) and REM \(322^{\mathrm{nd}}\) epochs show frequency mismatch, while the REM \(345^{\mathrm{th}}\) epoch shows non-respiratory movement contamination.
Thus, representing video with a single proxy may discard spatially localized evidence or introduce proxy-specific errors.

\begin{figure}[t]
    \centering
    \includegraphics[width=\linewidth]{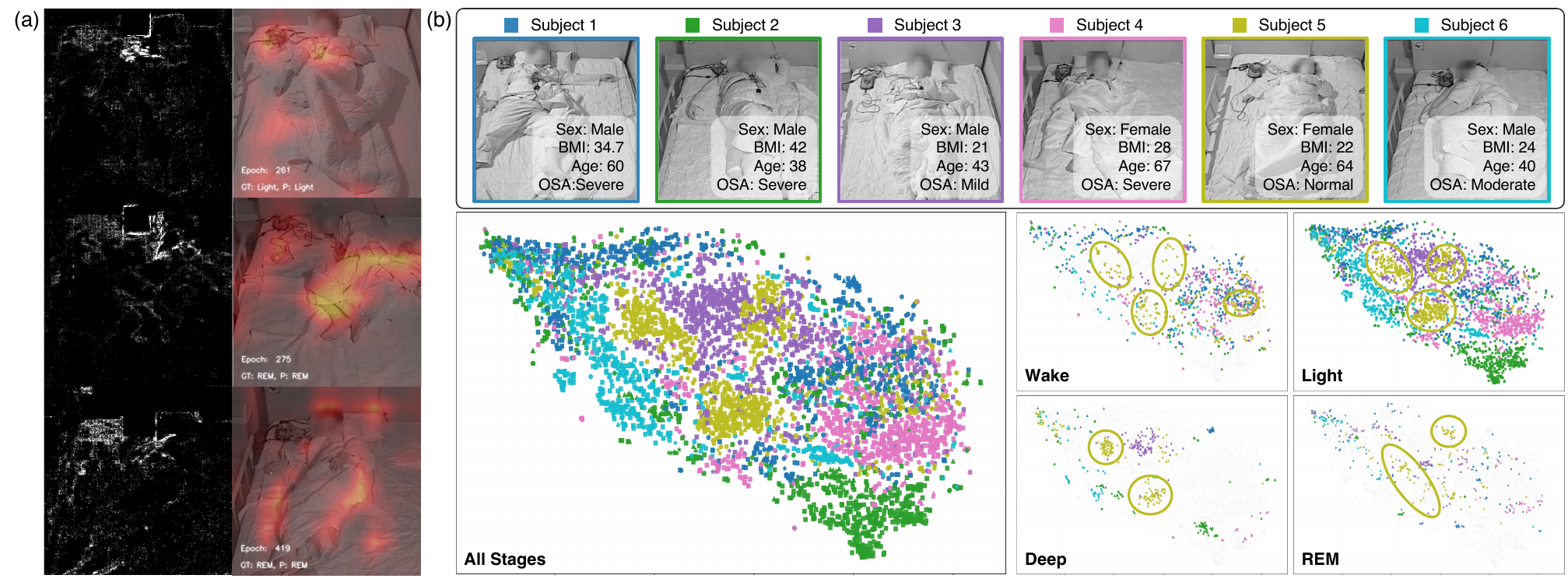}
    \caption{
    Baseline 3D CNN behavior on raw NIR sleep videos.
    (a) Frame-difference and spatial attention maps highlight static high-contrast regions rather than temporally coherent thoraco-abdominal motion.
    (b) t-SNE~\cite{maaten2008tsne} projection of epoch-level embeddings from six subjects shows subject-dominant clustering and within-stage subject dispersion.
    }
    \vspace{-2ex}
    \label{fig:design}
\end{figure}

Naively training a generic video encoder on raw NIR clips is also insufficient.
As shown in Figure~\ref{fig:design}(a), a baseline 3D CNN (MoViNet-A2~\cite{kondratyuk2021movinets}) attends to static high-contrast structures such as bedding folds, sensor cables, or limb boundaries, even when frame differences indicate motion around the thoraco-abdominal region.
Figure~\ref{fig:design}(b) shows that epoch-level embeddings cluster primarily by subject rather than by sleep stage, with substantial dispersion across subjects even within the same stage. 
This subject-dominant structure reflects differences in body shape, bedding, camera geometry, OSA severity, and baseline respiration-related motion.
Moreover, motion patterns can vary across the night even within the same subject and stage, reinforcing the need to interpret local video evidence within a full-night trajectory.

\begin{figure}[t]
  \centering
  \includegraphics[width=\linewidth]{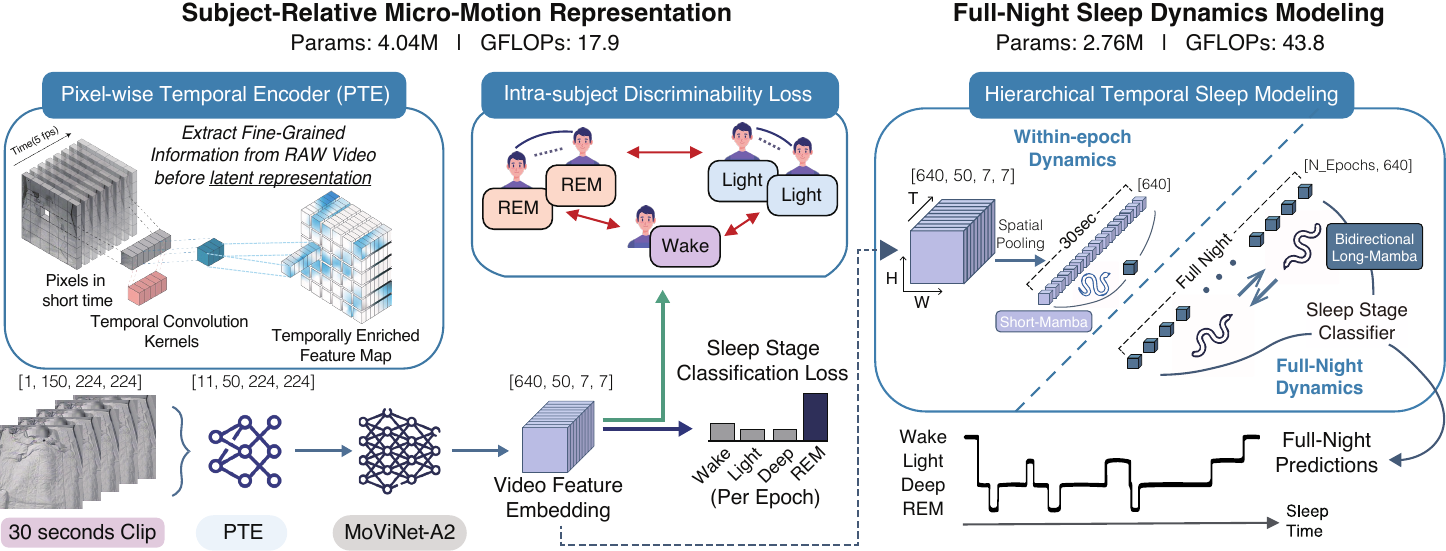}
  \caption{
  Overview of \proposal.
  }
  \label{fig:overview}
\end{figure}

Taken together, these analyses motivate learning directly from spatially localized video evidence rather than reducing each epoch to a reconstructed proxy or a generic raw-video embedding.
Since single-epoch visual evidence remains indirect and sleep stages evolve across the night, \proposal{} preserves weak local micro-motion before spatial aggregation, learns stage evidence relative to each subject's overnight baseline, and models sleep dynamics over the full night.
Figure~\ref{fig:overview} summarizes how \proposal{} implements these design goals.

%%%%%%%%%%%%%%%%%%%%%%%%%%%%%%%%%%%%%%%%%%%%%%%%%%%%%%
%%%%%%%%%%%%%%%%%%%%%%%%%%%%%%%%%%%%%%%%%%%%%%%%%%%%%%
\subsection{Subject-Relative Micro-Motion Representation}
\label{sec:micro_motion}

We first describe how \proposal{} forms epoch-level video representations before full-night modeling.
The representation module addresses the first two design goals by enriching each 30-second clip with pixel-wise temporal responses and using within-subject stage contrast to reduce subject-dominant structure during training.

%%%%%%%%%%%%%%%%%%%%%%%%%
\paragraph{Spatially anchored pre-spatial micro-motion encoding.}

Let $X_t \in \mathbb{R}^{C \times T \times H \times W}$ denote the video clip for epoch $t$, with channel, frame, height, and width dimensions. 
Standard video backbones may mix and downsample local regions before weak temporal evidence becomes separable. 
This is problematic for NIR sleep video because sleep-relevant motion can be highly localized and may be diluted unless temporal variation is first encoded at each pixel. 
To preserve dense pixel-level temporal traces over the epoch before spatial reduction, we introduce a Pixel-Wise Temporal Encoder (PTE):
\begin{equation}
\tilde{X}_t =
\operatorname{Proj}
\left(
\operatorname{Concat}
\left[
X_t,
M_{3}(X_t),
M_{5}(X_t)
\right]
\right),
\end{equation}
where $M_{3}(\cdot)$ and $M_{5}(\cdot)$ are learnable temporal encoding branches with kernel sizes 3 and 5, applied along the frame axis independently at each spatial location.
$\operatorname{Concat}(\cdot)$ denotes channel-wise concatenation with the original clip, and $\operatorname{Proj}(\cdot)$ is a shallow temporal projection layer that reduces the frame dimension via temporal striding before the video backbone.
PTE does not introduce a separate spatial stream; spatial mixing is deferred to the backbone.
Detailed architecture is provided in Appendix~\ref{appendix:implementation_details}.

PTE applies short temporal filters over the full 30-second epoch while preserving the spatial location of each temporal response.
Although each convolution uses a small kernel, stacking expands the effective receptive field to local sub-epoch intervals of roughly one to three seconds at 5\,fps.
Thus, PTE targets local motion traces rather than full-epoch or longer-range respiratory dynamics.
It exposes displacement-like variation, short oscillatory changes, and transient movement onsets before they are mixed with static appearance.
Unlike hand-crafted frame differences or late-fusion motion branches, PTE produces learnable, spatially localized micro-motion traces before backbone-level spatial mixing.
The enriched clip is then encoded by a compact MoViNet-A2 backbone, producing a spatiotemporal feature map \(F_t\).

%%%%%%%%%%%%%%%%%%%%%%%%%
\paragraph{Within-subject stage contrast.}
Even after motion-sensitive encoding, epoch features may reflect subject- and acquisition-specific factors rather than stage-related motion. 
We therefore contrast sleep stages within the same overnight recording.
For subject $s$, an anchor and a positive epoch are sampled from the same stage, while a negative epoch is sampled from a different stage of the same subject.
This within-subject sampling makes the comparison subject-relative, encouraging the representation to emphasize stage-related motion differences with respect to the subject's night-specific baseline.

We implement this objective as an Intra-Subject Discriminability Loss (ISDL).
Let $(x_{s,i}^a,x_{s,i}^p,x_{s,i}^n)$ denote the $i$-th anchor-positive-negative triplet of epoch features for subject $s$, and let $f(\cdot)$ be a projection head.
Given $N_{\mathrm{tri}}$ sampled triplets and margin $\alpha$, ISDL is
\begin{equation}
\mathcal{L}_{\text{ISDL}}
=
\frac{1}{N_{\mathrm{tri}}}
\sum_i
\left[
\|f(x_{s,i}^a)-f(x_{s,i}^p)\|_2^2
-
\|f(x_{s,i}^a)-f(x_{s,i}^n)\|_2^2
+
\alpha
\right]_+ .
\end{equation}
This follows a triplet-style margin objective~\cite{triplet}, but with within-subject sampling.
The representation objective combines the sleep-stage classification cross-entropy loss $\mathcal{L}_{\text{CE}}$ with ISDL:
\begin{equation}
\mathcal{L}_{\text{rep}} =
\mathcal{L}_{\text{CE}} + \lambda \mathcal{L}_{\text{ISDL}},
\end{equation}
where $\lambda$ controls the weight of the contrastive regularizer.

The key role of ISDL is not the margin loss itself, but the sampling semantics.
Unlike dataset-level contrastive objectives, ISDL forms comparisons within the same overnight recording, reducing variation from cross-subject and acquisition differences.

%%%%%%%%%%%%%%%%%%%%%%%%%%%%%%%%%%%%%%%%%%%%%%%%%%%%%%
%%%%%%%%%%%%%%%%%%%%%%%%%%%%%%%%%%%%%%%%%%%%%%%%%%%%%%
\subsection{Full-Night Sleep Dynamics Modeling}
\label{sec:sdm}

Although the representation module produces motion-sensitive features for each epoch, sleep staging is inherently sequential.
We introduce Sleep Dynamics Modeling (SDM) as a two-scale sequence module that first summarizes within-epoch motion evolution and then models the resulting epoch representations across the full night.
This design allows temporal modeling to operate on compact video-derived motion summaries rather than frame-level features or post-hoc label sequences.
We instantiate both temporal scales with Mamba~\cite{gu2023mamba}; implementation details are provided in Appendix~\ref{appendix:implementation_details}.

%%%%%%%%%%%%%%%%%%%%%%%%%
\paragraph{Within-epoch motion dynamics.}
The backbone output \(F_t\) retains spatial and temporal dimensions.
We first pool the spatial dimensions while preserving the temporal axis:
\begin{equation}
R_t = \operatorname{SpatialPool}(F_t).
\end{equation}
ShortMamba then summarizes the temporal evolution within the 30-second epoch into an epoch-level motion representation:
\begin{equation}
z_t = \operatorname{ShortMamba}(R_t).
\end{equation}
This step condenses within-epoch temporal dynamics, such as respiration-related periodicity and transient body movement, before full-night sequence modeling.

%%%%%%%%%%%%%%%%%%%%%%%%%
\paragraph{Full-night sleep dynamics.}
The epoch-level motion representations are arranged chronologically as
\(\mathcal{Z}=\{z_1,\ldots,z_N\}\) and processed by a bidirectional LongMamba module:
\begin{equation}
(s_1,\ldots,s_N)=\operatorname{LongMamba}(z_1,\ldots,z_N).
\end{equation}
The sleep-stage probability for epoch \(t\) is then
\begin{equation}
p(y_t \mid \mathcal{V})=\operatorname{softmax}(\operatorname{Classifier}(s_t)).
\end{equation}

SDM operates on video-derived epoch representations before classification, rather than smoothing predicted labels after classification.
By relating local motion summaries to the subject's full-night trajectory, SDM helps calibrate subject-specific motion baselines, maintain coherent sleep progression, and reduce abrupt epoch-wise fluctuations.

\section{Experiments}
\label{sec:experiments}

%%%%%%%%%%%%%%%%%%%%%%%%%%%%%%%%%%%%%%%%%%%%%%%%%%%%%%
%%%%%%%%%%%%%%%%%%%%%%%%%%%%%%%%%%%%%%%%%%%%%%%%%%%%%%
\subsection{Experimental Setup}
\label{sec:experimentalSetup}

\paragraph{Dataset.}
We evaluate on the Korea Video Sleep Study (KVSS) dataset~\cite{aihub_kvss,choi2026non}, using 475 overnight NIR sleep recordings ($\sim$3,250 hours) paired with synchronized PSG.
Sleep stages were annotated by certified technicians according to AASM guidelines~\cite{AASM}.
The cohort includes healthy and sleep-disordered individuals with diverse age, BMI, sex, and OSA severity.
Videos were captured at 5\,fps using a wall-mounted NIR camera under routine overnight sleep-study conditions, including far-field viewpoints, bedding occlusion, weak motion, and face anonymization.
Subjects were split into disjoint train/validation/test sets of 335/70/70 using stratified sampling over sex, age, and OSA severity.
Additional acquisition details and cohort characteristics are provided in Appendix~\ref{appendix:kvss}.

\paragraph{Implementation.}
We implement \proposal{} in PyTorch and train on four NVIDIA RTX A6000 GPUs.
The model is trained with five AASM sleep-stage labels (Wake, N1, N2, N3, REM), and N1/N2 are merged into Light sleep for the reported four-class evaluation.
We report accuracy, Cohen's kappa ($\kappa$), and macro-F1 (MF1).
MF1 is emphasized because sleep stages are imbalanced and minority stages such as Deep and REM are clinically important.
Architecture, preprocessing, optimization, and hyperparameter details are provided in Appendix~\ref{appendix:implementation_details}. The submitted anonymized implementation will be made publicly available upon acceptance.

%%%%%%%%%%%%%%%%%%%%%%%%%%%%%%%%%%%%%%%%%%%%%%%%%%%%%%
%%%%%%%%%%%%%%%%%%%%%%%%%%%%%%%%%%%%%%%%%%%%%%%%%%%%%%
\subsection{Overall Performance}

\begin{table}[t]
\centering
\caption{
Performance comparison of video-based sleep staging methods.
Results for non-KVSS datasets are taken from their respective papers, while KVSS results are from our own implementation.
\textit{Oracle} indicates experiments using ground-truth physiological signals from contact sensors.
}
\label{tab:performance_comparison}
\small
\setlength{\tabcolsep}{6.5pt}
\renewcommand{\arraystretch}{.6}

\begin{tabular}{l c c c c c c}
\toprule
\textbf{Method} & 
\textbf{Input / Proxy} & 
\textbf{Dataset} & 
\textbf{\# of} \textbf{Data} & 
$\mathbf{Acc}$ & 
$\boldsymbol{\kappa}$ & 
$\mathbf{MF1}$ \\
\midrule

van Meulen et al.~\cite{van2023contactless}
& HRV
& HealthBed
& 46 
& 0.68 & 0.49 & 0.63 \\

Video-PSG~\cite{wang2024video} 
& HR+HRV+BR+M
& Shenzhen
& 20 
& 0.73 & 0.62 & 0.74 \\

SleepVST~\cite{carter2024sleepvst} 
& BW+CW+M
& Oxford
& 50 
& 0.79 & 0.71 & 0.78 \\

\midrule

\multirow{2}{*}{SleepVST~\cite{carter2024sleepvst}}
& BW*+CW*+M
& \multirow{2}{*}{KVSS}
& \multirow{2}{*}{475}
& 0.78 & 0.63 & 0.73 \\
& BW+CW*+M
& 
& 
& 0.61 & 0.23 & 0.37 \\

\midrule

MoViNet-A2~\cite{kondratyuk2021movinets}
& Video only
& KVSS
& 475
& 0.60 & 0.35 & 0.50 \\

MoViNet-A2~\cite{kondratyuk2021movinets} + PTE
& Video only
& KVSS
& 475
& 0.62 & 0.40 & 0.55 \\

MoViNet-A5~\cite{kondratyuk2021movinets} + PTE
& Video only
& KVSS
& 475
& 0.60 & 0.38 & 0.53 \\

Video Swin-Tiny~\cite{swin} + PTE
& Video only
& KVSS
& 475
& 0.51 & 0.27 & 0.46 \\

\midrule

\textbf{\proposal}
& \textbf{Video only}
& KVSS 
& 475
& \textbf{0.80} & \textbf{0.68} & \textbf{0.78} \\

\bottomrule
\end{tabular}
\vspace{0.2ex}

{\scriptsize\itshape
Acc: accuracy;
$\kappa$: Cohen's kappa;
MF1: macro F1-score.
BR: breathing rate; HR: heart rate; HRV: heart-rate variability; \\
BW: breathing waveform; CW: cardiac waveform; M: motion feature.
Asterisks (*) indicate waveforms from PSG sensors (oracle setting). \\
Rows marked ``+PTE'' include only the Pixel-Wise Temporal Encoder, excluding ISDL and SDM.
\par}
\vspace{-1.5ex}
\end{table}

Table~\ref{tab:performance_comparison} compares \proposal{} with prior video-based sleep staging methods.
Results from non-KVSS datasets are included for reference, as datasets differ in acquisition protocols, cohort size, camera setup, and label distribution.
For controlled comparisons on KVSS, we include SleepVST~\cite{carter2024sleepvst}, a representative proxy-first method, and video-only backbone variants.

Reproducing the fully video-based SleepVST pipeline on KVSS is not feasible because its video-to-waveform extraction module is not publicly released, and face anonymization and bedding occlusion make rPPG and respiratory reconstruction unreliable.
We therefore evaluate two oracle-input variants: a partial-oracle setting with contact-sensor cardiac waveforms and a full-oracle setting with both contact-sensor respiratory and cardiac waveforms.
These variants are oracle-input versions of one representative proxy-first architecture, not universal upper bounds for physiological-signal-based sleep staging; details are provided in Appendix~\ref{appendix:sleepvst}.

Video-only backbone variants perform substantially below \proposal{}.
Although PTE improves the vanilla MoViNet-A2 baseline, the gap to the full \proposal{} remains large, indicating that pre-spatial micro-motion encoding alone is insufficient without within-subject stage contrast and full-night dynamics.
MoViNet-A5+PTE does not improve over MoViNet-A2+PTE despite its larger size, and Video Swin-Tiny+PTE performs worse, suggesting that simply scaling or replacing the backbone is insufficient for NIR sleep video.
MoViNet-A2 is used in \proposal{} because it provides the best accuracy-efficiency tradeoff among the evaluated variants.

On KVSS, \proposal{} achieves 0.80 accuracy, 0.68 $\kappa$, and 0.78 MF1.
Compared with the full-oracle SleepVST variant, \proposal{} improves MF1 from 0.73 to 0.78.
This result supports NIR video as an independently informative and complementary modality for PSG-defined sleep-stage estimation, while PSG remains the clinical reference for defining the labels.
Appendix~\ref{appendix:stratified_perf} further reports stratified per-subject analyses across recording conditions and subject characteristics, including occlusion, camera angle, illumination, sex, age, BMI, and AHI groups.

%%%%%%%%%%%%%%%%%%%%%%%%%%%%%%%%%%%%%%%%%%%%%%%%%%%%%%
%%%%%%%%%%%%%%%%%%%%%%%%%%%%%%%%%%%%%%%%%%%%%%%%%%%%%%
\subsection{Ablation Study}

%%%%%%%%%%%%%%%%%%%%%%%%%%%
\paragraph{Effect of PTE and ISDL.}

\begin{wraptable}{r}{0.5\textwidth}
\vspace{-3ex}
\centering
\caption{Ablation study of key components on KVSS. Exp.1 corresponds to the vanilla MoViNet-A2. \cmark{} indicates that each component is applied.}
\label{tab:ablation_study}
\vspace{-0.4ex}
\fontsize{9}{9}\selectfont
\setlength{\tabcolsep}{5pt}
\renewcommand{\arraystretch}{0.8}
\begin{tabular}{cccccc}
\toprule
\multirow{2}{*}{\textbf{Exp.\#}}
  & \multicolumn{3}{c}{\textbf{Method}}
  & \multirow{2}{*}{\textbf{Acc.}}
  & \multirow{2}{*}{\textbf{MF1}} \\
\cmidrule(lr){2-4}
  & \textbf{PTE} & \textbf{ISDL} & \textbf{SDM} &  &  \\
\midrule
1 & \xmark & \xmark & \xmark & 0.60 & 0.50 \\
2 & \cmark & \xmark & \xmark & 0.62 & 0.55 \\
3 & \cmark & \cmark & \xmark & 0.60 & 0.56 \\
4 & \xmark & \xmark & \cmark & 0.75 & 0.67 \\
5 & \cmark & \xmark & \cmark & 0.77 & 0.72 \\
6 & \cmark & \cmark & \cmark & \textbf{0.80} & \textbf{0.78} \\
\bottomrule
\end{tabular}
\vspace{-3.5ex}
\end{wraptable}

Table~\ref{tab:ablation_study} shows the contribution of each component, starting with the representation modules without SDM.
The baseline (Exp.~1) achieves 0.60 accuracy and 0.50 MF1.
Adding PTE (Exp.~2) improves accuracy to 0.62 and MF1 to 0.55, indicating that pre-spatial micro-motion encoding helps preserve weak local temporal cues before spatial processing.
Adding ISDL (Exp.~3) further improves MF1 to 0.56, although accuracy remains comparable, suggesting that within-subject stage contrast improves stage-wise separability under subject-specific variation.
Overall, PTE and ISDL improve representation quality from complementary perspectives: PTE enhances motion sensitivity, while ISDL regularizes the embedding space to reduce subject-dependent bias.

%%%%%%%%%%%%%%%%%%%%%%%%%%%
\paragraph{Effect of SDM.}

Adding SDM to the vanilla backbone (Exp.~4) improves MF1 from 0.50 to 0.67, showing the importance of full-night temporal context.
However, the benefit of SDM depends strongly on the quality of the epoch-level representation.
With the same SDM module, adding PTE (Exp.~5) increases MF1 from 0.67 to 0.72, and adding both PTE and ISDL (Exp.~6) further increases MF1 to 0.78.
Thus, the representation and dynamics modules are complementary: PTE and ISDL make the epoch-level video evidence more informative and subject-relative, while SDM organizes this evidence into a coherent full-night sleep-stage trajectory.

\begin{figure}[t]
    \centering
    \includegraphics[width=\linewidth]{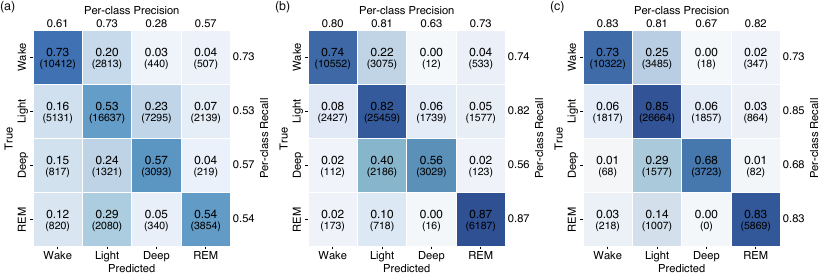}
    \vspace{-1.5ex}
    \caption{
    Confusion matrices on the KVSS test set for four-class sleep staging.
    (a) Without SDM.
    (b) SDM with LongMamba only.
    (c) Full SDM with ShortMamba and LongMamba.
    }
    \label{fig:confusion_matrix}
    \vspace{-0.5ex}
\end{figure}

Figure~\ref{fig:confusion_matrix} shows how SDM changes class-wise prediction behavior beyond the aggregate scores in Table~\ref{tab:ablation_study}.
Without SDM (a), predictions are broadly less balanced, with substantial confusion between Light and neighboring stages and low precision for Deep sleep.
Adding LongMamba only (b) improves the overall prediction structure by incorporating full-night temporal context, increasing recall for Light and REM and improving precision for Wake and Deep.
This indicates that full-night context helps organize epoch-level predictions into a more consistent sleep-stage trajectory.
The full SDM (c) further incorporates within-epoch motion summarization.
Compared with LongMamba only, it improves Deep precision and recall from 0.63/0.56 to 0.67/0.68 and increases REM precision from 0.73 to 0.82, although REM recall slightly decreases from 0.87 to 0.83.
Together, these results support the two-scale design, where LongMamba provides trajectory-level context and ShortMamba refines the epoch-level motion evidence used by the full-night model.
Additional experiments of SDM design choices are provided in Appendix~\ref{appendix:ablation_sdm}, including temporal context-length sweeps and comparisons with alternative sequence models.

%%%%%%%%%%%%%%%%%%%%%%%%%%%
\paragraph{Full-night prediction behavior.}

\begin{figure}[t]
    \centering
    \includegraphics[width=\linewidth]{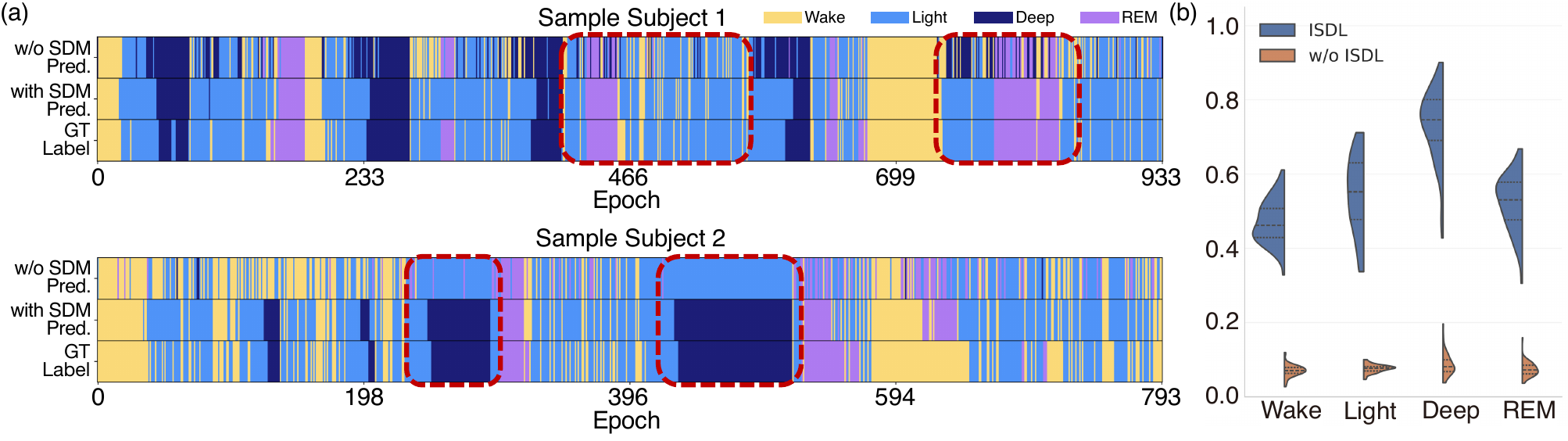}
    \caption{
    Full-night prediction behavior and temporal coherence. 
    (a) Full-night predictions. Red boxes highlight representative segments. (b) Temporal coherence.
    }
    \vspace{-2ex}
    \label{fig:full-night}
\end{figure}

Figure~\ref{fig:full-night}(a) illustrates how SDM affects complete overnight predictions.
For Subject~1, predictions without SDM frequently oscillate between adjacent stages in the highlighted regions, whereas SDM suppresses abrupt fluctuations and preserves smoother stage trajectories.
For Subject~2, predictions without SDM fail to recover sustained Deep sleep segments, while SDM accumulates temporally consistent evidence and restores continuous Deep predictions.
These examples show that full-night dynamics not only smooths noisy local predictions, but also helps recover sustained sleep structure when single-epoch visual evidence is weak.

Figure~\ref{fig:full-night}(b) further shows how ISDL interacts with full-night dynamics by improving the temporal coherence of epoch-level representations.
We measure coherence as the cosine similarity between consecutive epoch embeddings, \(s_t=\cos(\mathbf{f}_t,\mathbf{f}_{t+1})\).
ISDL increases this coherence across stages, indicating that within-subject stage contrast produces smoother feature trajectories within each subject.
These more stable epoch representations provide better inputs for SDM, explaining why combining PTE, ISDL, and SDM yields the strongest performance.
Together with Table~\ref{tab:ablation_study} and Figure~\ref{fig:confusion_matrix}, these results show that micro-motion representation, subject-relative contrast, and full-night temporal modeling provide complementary benefits for video-only sleep staging.

%%%%%%%%%%%%%%%%%%%%%%%%%%%%%%%%%%%%%%%%%%%%%%%%%%%%%%
%%%%%%%%%%%%%%%%%%%%%%%%%%%%%%%%%%%%%%%%%%%%%%%%%%%%%%
\subsection{Interpretation Analysis}

To inspect the visual evidence used by \proposal{}, we apply Sleep Video EigenCAM (SVECAM), a visualization method adapted from EigenCAM~\cite{eigencam} for 3D sleep-video features.
SVECAM produces class-agnostic spatial saliency maps and temporal saliency curves, allowing us to inspect where and when the model attends within each epoch.
Implementation details are provided in Appendix~\ref{appendix:svecam}.

\begin{figure}[t]
    \centering
    \includegraphics[width=\linewidth]{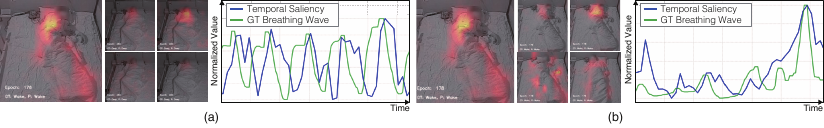}
    \vspace{-2ex}
    \caption{
    SVECAM visualizations for representative NIR sleep-video epochs.
    (a) Stable Deep sleep, showing thoraco-abdominal saliency and temporal saliency aligned with breathing.
    (b) Brief arousals and position change, showing both respiration-related motion and larger body movement.
    }
    \label{fig:eigencam}
    \vspace{-1em}
\end{figure}

Figure~\ref{fig:eigencam} shows two representative cases.
During stable Deep sleep (a), saliency concentrates around the thoraco-abdominal region, and temporal saliency is aligned with respiratory rhythm despite the absence of explicit respiratory reconstruction.
During arousals and position changes (b), the model still captures respiration-related motion but also attends to larger body movements.
These visualizations suggest that \proposal{} uses physiologically meaningful motion evidence encoded in NIR video, without explicitly reconstructing predefined physiological proxies.

The observed saliency patterns also help explain the difficulty of single-epoch video-only staging.
Stable Deep sleep often depends on subtle periodic motion with little gross movement, whereas arousal-related epochs can produce broader and more transient saliency.
Light sleep and REM can remain visually heterogeneous, often sharing low-motion or transition-heavy patterns with neighboring stages.
This ambiguity supports the need for both subject-relative micro-motion representation and full-night sleep dynamics modeling.
\section{Conclusion}
\label{sec:conclusion}

We presented \proposal{}, a framework for PSG-defined sleep staging from near-infrared (NIR) video that combines subject-relative micro-motion learning with full-night sleep dynamics modeling.
\proposal{} preserves localized temporal motion before spatial reduction, learns stage evidence relative to each subject's overnight baseline, and organizes video-derived epoch evidence into coherent full-night sleep-stage trajectories.
On 475 overnight NIR recordings, \proposal{} achieves strong four-class staging performance, and ablations show that micro-motion representation, within-subject stage contrast, and sleep dynamics provide complementary benefits.
Sleep Video EigenCAM further suggests that the model uses physiologically meaningful visual motion, including thoraco-abdominal periodicity and broader body movements associated with arousals and position changes.
These results support NIR video as an independently informative and complementary modality for contactless sleep-stage estimation under PSG-defined labels.
Broader validation across additional clinical sites and home environments remains important, and future work may explore integration with physiological signals or wearable modalities for more comprehensive sleep monitoring.

\bibliographystyle{plain}
\bibliography{main}

%%%%%%%%%%%%%%%%%%%%%%%%%%%%%%%%%%%%%%%%%%%%%%%%%%%%%%%%%%%%

\appendix
\clearpage
\renewcommand{\thesection}{\Alph{section}}
\setcounter{section}{0}
\counterwithin{figure}{section}
\counterwithin{table}{section}

\section*{Appendix}
\label{sec:appendix}
\addcontentsline{toc}{section}{Appendix}

%%%%%%%%%%%%%%%%%%%%%%%%%
%%%%%%%%%%%%%%%%%%%%%%%%%
\section{KVSS Dataset}
\label{appendix:kvss}

%%%%%%%%%%%%%%%%%%%%%%%%%
\subsection{Dataset Composition}

\begin{figure*}[ht]
    \centering
    \includegraphics[width=.8\linewidth]{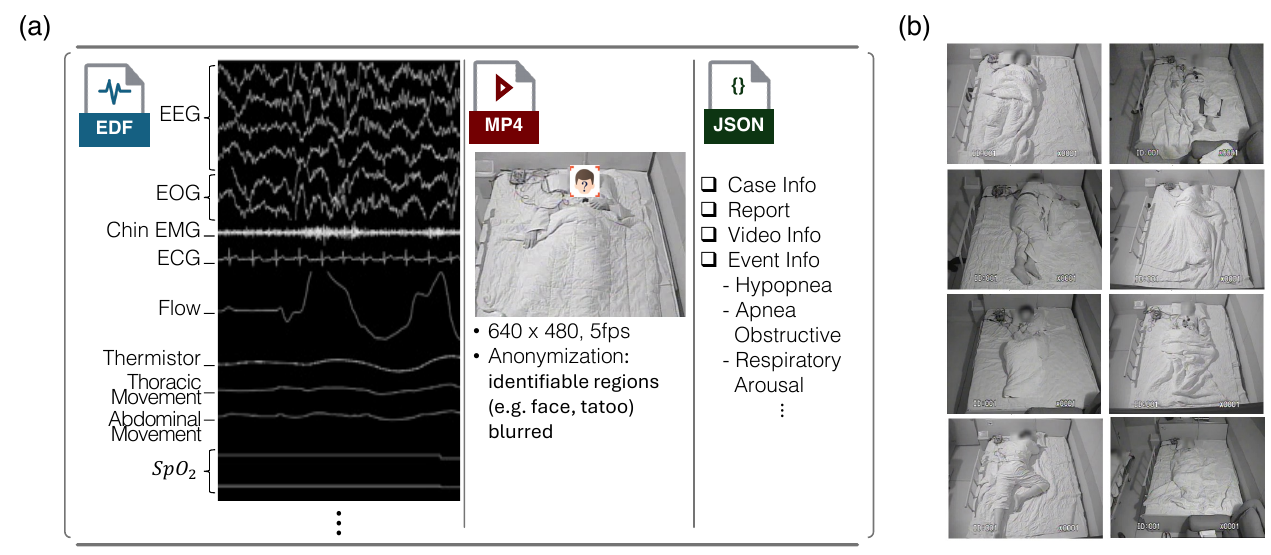}
    \caption{
    Composition of the KVSS dataset.
    (a) Overnight PSG acquisition with synchronized multimodal signals, anonymized NIR video, and expert sleep-stage annotations.
    (b) Representative NIR frames showing variation in subjects, body positions, bedding conditions, and room layouts.
    }
    \vspace{-1ex}
    \label{fig:supple_kvss}
\end{figure*}

Figure~\ref{fig:supple_kvss} summarizes the data composition and acquisition pipeline of the Korea Video Sleep Study (KVSS)~\cite{aihub_kvss,choi2026non}.
As shown in Figure~\ref{fig:supple_kvss}(a), the Hospital A subset contains 475 overnight NIR sleep recordings (\(\sim\)3,250 hours) paired with synchronized PSG and structured clinical metadata.
PSG signals include EEG, EOG, EMG, ECG, airflow, thoracic and abdominal respiratory effort, and blood oxygen saturation (SpO$_2$), from which sleep stages were annotated by certified technicians following AASM guidelines~\cite{AASM}.
NIR videos were captured at 640\,$\times$\,480 resolution and 5\,fps using a wall-mounted camera positioned approximately 3.5\,m from the subject with an oblique view.
Identifiable regions, including faces and tattoos, were blurred to protect privacy while preserving motion cues relevant to sleep analysis.

The recordings were collected during routine overnight sleep studies with a fixed far-field viewpoint, where subjects were typically covered by bedding.
Figure~\ref{fig:supple_kvss}(b) illustrates substantial variation in body position, bedding, occlusion, and room layout.
Compared with closer-range or more controlled camera setups, these conditions attenuate visible motion, increase occlusion, and introduce inter-subject variability, making KVSS a challenging testbed for video-based sleep staging.
For all experiments, we split the 475 subjects into disjoint training, validation, and test sets of 335/70/70 using stratified sampling over sex, age, and OSA severity.

%%%%%%%%%%%%%%%%%%%%%%%%%
\subsection{Demographic and Clinical Characteristics}

\begin{figure*}[ht]
    \centering
    \includegraphics[width=.9\linewidth]{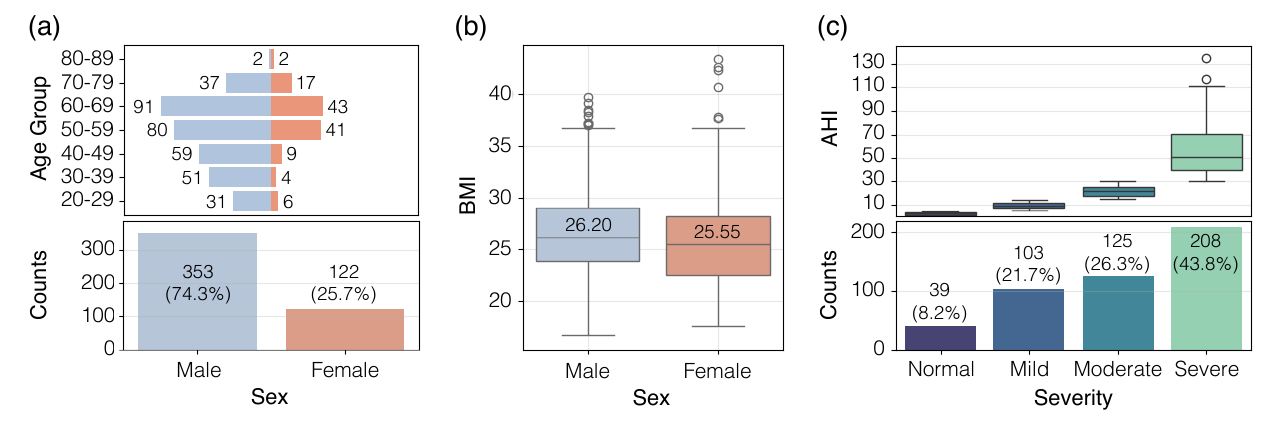}
    \caption{
    Demographic and clinical characteristics of the Hospital~A cohort.
    (a) Sex and age. 
    (b) BMI by sex. 
    (c) OSA severity and AHI.
    }
    \label{fig:supple_kvss_demo}
\end{figure*}

Figure~\ref{fig:supple_kvss_demo} summarizes the demographic and clinical characteristics of the 475 subjects from Hospital~A.
The cohort spans a broad age range and includes 353 males (74.3\%) and 122 females (25.7\%), reflecting the composition of patients undergoing clinical sleep testing.
BMI distributions show substantial inter-individual variability, with median BMI values of 26.20 for males and 25.55 for females.
The cohort also covers a wide OSA severity spectrum, with 39 normal (8.2\%), 103 mild (21.7\%), 125 moderate (26.3\%), and 208 severe cases (43.8\%) according to AHI.

Together, Figures~\ref{fig:supple_kvss} and~\ref{fig:supple_kvss_demo} show that the KVSS Hospital~A cohort is diverse in demographics, body type, clinical severity, and video acquisition conditions.
This diversity supports evaluation under heterogeneous sleep populations and realistic clinical recording conditions.

%%%%%%%%%%%%%%%%%%%%%%%%%
%%%%%%%%%%%%%%%%%%%%%%%%%
\section{\proposal{} Architecture and Experimental Details}
\label{appendix:implementation_details}

\subsection{\proposal{} Architecture}

\subsubsection{Subject-Relative Micro-Motion Representation.}

\begin{figure}[t]
    \centering
    \includegraphics[width=.8\linewidth]{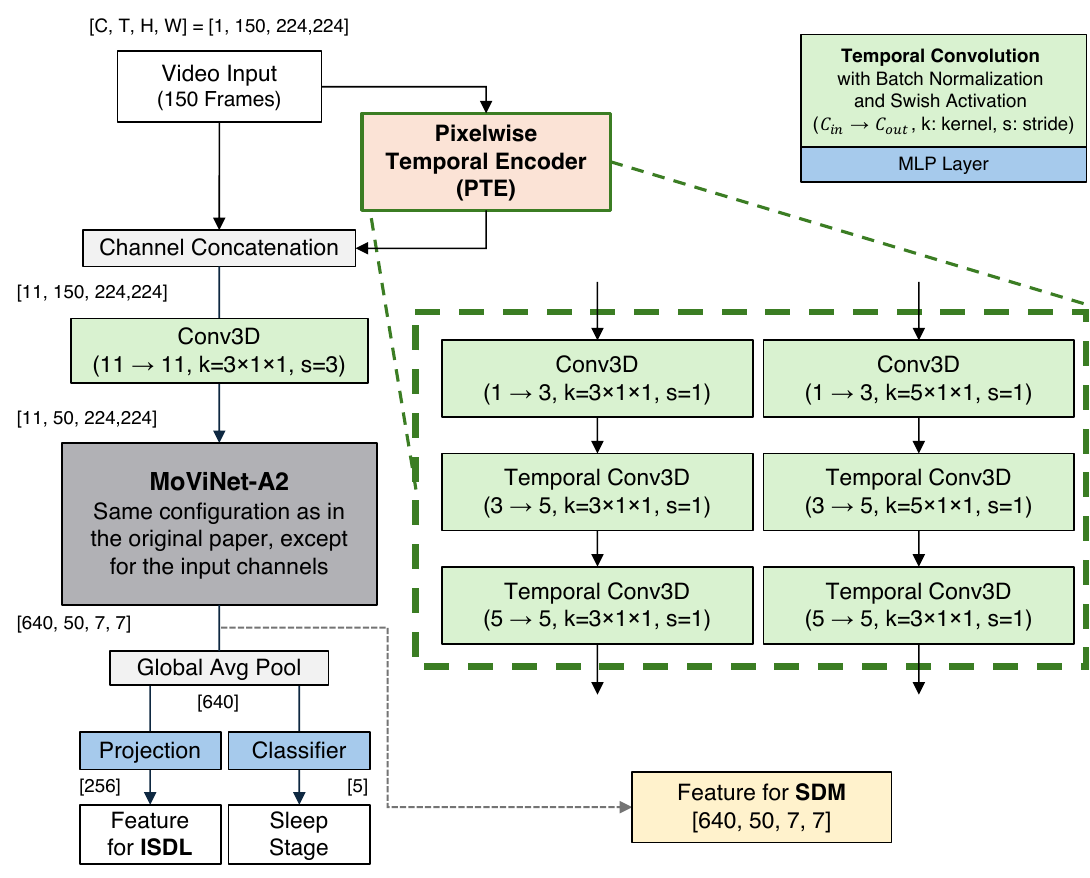}
    \caption{
    Architecture of the subject-relative micro-motion representation module.
    PTE enriches each 30-second epoch with pixel-wise temporal responses, which are concatenated with the original video before strided temporal projection and MoViNet-A2 encoding.
    GAP features are used for epoch-level classification and ISDL, while pre-GAP feature maps are passed to SDM.
    }
    \label{fig:supple_stage1}
\end{figure}

Figure~\ref{fig:supple_stage1} illustrates the detailed architecture of the subject-relative micro-motion representation module.
Each 30-second epoch, comprising 150 frames at 5\,fps, is first processed by the Pixel-Wise Temporal Encoder (PTE).
PTE consists of temporal-only convolution branches applied independently at each spatial location, without spatial mixing.
The branches use stacked \(k\times1\times1\) temporal convolutions with short kernels, producing local temporal responses that are concatenated with the original video along the channel dimension.
A strided \(3\times1\times1\) temporal projection then reduces the temporal resolution before MoViNet-A2 encoding.

Although the individual kernels are short, stacking expands the effective temporal receptive field.
Including the temporal projection, the two PTE branches cover local neighborhoods of up to 9 and 13 frames, corresponding to about 1.8\,s and 2.6\,s at 5\,fps.
Thus, PTE is designed to expose local sub-epoch motion traces before spatial mixing, while longer-range within-epoch dynamics are handled by the sleep dynamics modeling module.
All MoViNet-A2 settings follow the original specification~\cite{kondratyuk2021movinets}, except that the first convolutional layer is modified to accommodate the increased number of input channels after concatenation.

The MoViNet-A2 backbone produces spatiotemporal feature maps.
Global average pooling (GAP) yields epoch-level features used for sleep-stage classification and the Intra-Subject Discriminability Loss (ISDL).
The pre-GAP feature maps are retained as input to the sleep dynamics modeling module.

\begin{figure}[t]
    \centering
    \includegraphics[width=.85\linewidth]{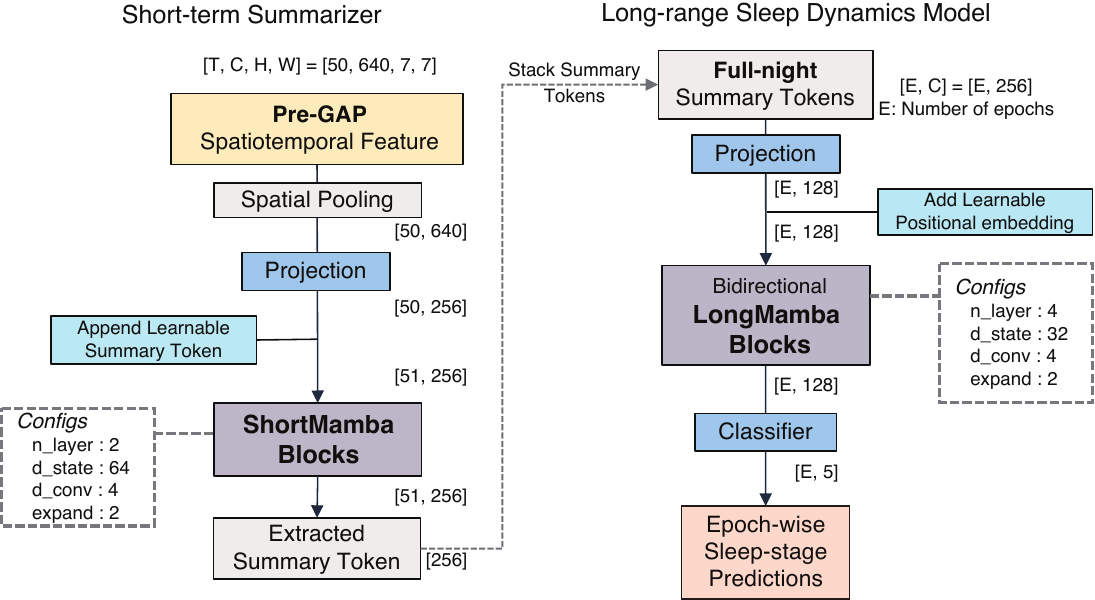}
    \caption{
    Architecture of the sleep dynamics modeling module.
    ShortMamba summarizes each epoch by spatially pooling pre-GAP feature maps, appending a learnable summary token, and extracting the summary-token output.
    Bidirectional LongMamba models the stacked epoch tokens over the full night with positional embeddings and outputs stage predictions for all epochs.
    }
    \label{fig:supple_stage2}
\end{figure}

%%%%%%%%%%%%%%%%%%%%%%%%%
\subsubsection{Full-Night Sleep Dynamics Modeling.}

Figure~\ref{fig:supple_stage2} illustrates the architecture of Sleep Dynamics Modeling (SDM), which operates on the pre-GAP spatiotemporal feature maps from the representation module.
SDM has two temporal scales: ShortMamba summarizes motion evolution within each 30-second epoch, and bidirectional LongMamba models the resulting epoch tokens across the full night.

For each epoch, pre-GAP feature maps are spatially pooled into a temporal sequence with 50 temporal positions.
After projection, a learnable summary token is appended to the sequence.
ShortMamba models within-epoch temporal dynamics, and the output corresponding to the summary token is used as the compact epoch-level motion representation.

The epoch-level tokens are stacked chronologically across the full night.
Each token is augmented with a learnable positional embedding, and the resulting sequence is processed by bidirectional LongMamba.
The updated tokens are fed into a shared classifier to produce sleep-stage predictions for all epochs.
This two-scale design captures within-epoch motion evolution and full-night sleep progression while avoiding post-hoc smoothing of predicted labels.
The Mamba implementation enables linear-time modeling of the full-night epoch sequence.

%%%%%%%%%%%%%%%%%%%%%%%%%
\subsection{\proposal{} Training Details}

\paragraph{Video Preprocessing.}
All NIR sleep videos in KVSS were stored as MP4 files at 640\,$\times$\,480 resolution and 5\,fps.
For each night, frames were cropped to a 400\,$\times$\,400 region centered on the bed to retain body motion while removing irrelevant background, and then resized to 224\,$\times$\,224 to reduce computational cost.
The continuous video stream was segmented into 30\,s epochs aligned with sleep-stage annotations, so that each epoch contained 150 frames.
For epochs with fewer than 150 frames, such as recording boundaries, we applied last-frame padding to enforce a fixed temporal length.
For efficient I/O during training, all epochs from each subject were stored as fixed-size tensors in a single HDF5 file and accessed directly by the data loader.

\paragraph{Subject-Relative Micro-Motion Representation Training.}
The MoViNet-A2 backbone in the subject-relative micro-motion representation module is initialized from a publicly available Kinetics-600-pretrained checkpoint~\cite{kinetics,kondratyuk2021movinets} and fine-tuned on KVSS.
Training is performed on four NVIDIA A6000 GPUs using weighted random sampling.
Each optimizer epoch samples approximately 9\% of the full training set.
The effective batch size is 8 triplet samples, corresponding to 24 sleep epochs per optimization step because each sample contains an anchor, positive, and negative epoch for within-subject contrast.
We use AdamW with a learning rate of \(10^{-3}\).
The module is optimized with the sum of the epoch-level cross-entropy loss and ISDL.
For ISDL triplet construction, we use class-balanced anchor sampling to reduce the effect of sleep-stage imbalance. 
For each sampled subject, an anchor stage is first sampled uniformly from the five AASM stages with valid candidate epochs for that subject. 
An anchor epoch is then sampled from the selected subject and stage. 
The positive epoch is sampled from the same subject and stage, and the negative epoch is sampled from the same subject but a different stage. 
Positive and negative epochs are selected uniformly from their respective candidate pools. 
We apply the following on-the-fly augmentations: normalization with dataset mean and standard deviation, random brightness jitter (\(\pm 0.2\)), random contrast jitter (\(\pm 0.2\)), horizontal flip with probability 0.5, random rotation up to \(\pm 30^\circ\), and random scaling uniformly sampled from \([0.8, 1.1]\).
The model is trained with five AASM stage labels, Wake, N1, N2, N3, and REM.
For four-class evaluation, N1 and N2 are merged into Light sleep.

\paragraph{Full-Night Sleep Dynamics Modeling Training.}
After training the subject-relative micro-motion representation module, we extract and store the pre-GAP spatiotemporal feature maps for all epochs.
These saved features are used as input to the sleep dynamics modeling module.
SDM is trained on four NVIDIA A6000 GPUs using subject-level sequences, where each data point corresponds to one full-night recording.
To handle variable-length nights without padding, sequences are processed individually and gradient accumulation is used to obtain an effective batch size of 64 subject-level sequences.
We use AdamW with a learning rate of \(10^{-4}\).

%%%%%%%%%%%%%%%%%%%%%%%%%
\subsection{Evaluation Metrics}

We evaluate four-class sleep staging performance using accuracy, macro-F1, and Cohen's \(\kappa\). 
For completeness, we summarize the definitions used in the main paper.

\paragraph{Accuracy.}
Accuracy measures the proportion of correctly classified epochs:
\[
\mathrm{Acc} = \frac{1}{N}\sum_{i=1}^{N}\mathbb{I}(\hat{y}_i = y_i),
\]
where \(N\) is the total number of evaluated epochs.

\paragraph{Macro-F1 (MF1).}
For each class \(c\), precision, recall, and F1-score are computed as
\[
\mathrm{Precision}_c = \frac{TP_c}{TP_c + FP_c}, \qquad
\mathrm{Recall}_c = \frac{TP_c}{TP_c + FN_c},
\]
\[
\mathrm{F1}_c =
2 \cdot
\frac{\mathrm{Precision}_c \cdot \mathrm{Recall}_c}
{\mathrm{Precision}_c + \mathrm{Recall}_c}.
\]
Macro-F1 is the unweighted average across classes:
\[
\mathrm{MF1} = \frac{1}{C}\sum_{c=1}^{C}\mathrm{F1}_c,
\]
where \(C\) is the number of sleep-stage classes. 
This metric assigns equal weight to each stage and is therefore useful under class imbalance.

\paragraph{Cohen's Kappa (\(\kappa\)).}
\(\kappa\) measures agreement between predictions and ground truth while accounting for chance agreement:
\[
\kappa = \frac{p_o - p_e}{1 - p_e},
\]
where \(p_o\) is the observed agreement and \(p_e\) is the expected agreement under random labeling:
\[
p_e = \sum_{c=1}^{C} 
\frac{n_{c,\mathrm{pred}} \cdot n_{c,\mathrm{true}}}{N^2}.
\]
Here, \(n_{c,\mathrm{pred}}\) and \(n_{c,\mathrm{true}}\) denote the number of predicted and true labels for class \(c\).

\paragraph{Per-subject accuracy for stratified analysis.}
For robustness analyses across recording conditions and subject characteristics, we compute accuracy separately for each subject:
\[
\mathrm{Acc}_s =
\frac{1}{N_s}\sum_{i=1}^{N_s}\mathbb{I}(\hat{y}_{s,i}=y_{s,i}),
\]
where \(N_s\) is the number of epochs for subject \(s\).
For a subgroup \(\mathcal{G}\), we report the mean per-subject accuracy:
\[
\mathrm{Acc}_{\mathcal{G}} =
\frac{1}{|\mathcal{G}|}\sum_{s\in\mathcal{G}}\mathrm{Acc}_s.
\]
This assigns equal weight to each subject within a subgroup, regardless of recording length.

%%%%%%%%%%%%%%%%%%%%%%%%%
%%%%%%%%%%%%%%%%%%%%%%%%%
\section{SleepVST Re-implementation}
\label{appendix:sleepvst}

Following the original SleepVST setup~\cite{carter2024sleepvst}, we pre-train the model on the SHHS~\cite{shhs} and MESA~\cite{mesa} contact-sensor datasets for four-class sleep staging (Wake, N1/N2, N3, REM).

%%%%%%%%%%%%%%%%%%%%%%%%%
\subsection{Datasets for SleepVST Training}

\paragraph{Sleep Heart Health Study (SHHS)~\cite{shhs}.}
SHHS is a large multicenter cohort designed to study the cardiovascular and respiratory consequences of sleep-disordered breathing in adults.
It includes overnight home polysomnography from more than 6{,}400 participants, together with demographic, anthropometric, and clinical questionnaires.
PSG recordings include EEG, EOG, EMG, airflow, respiratory effort, and SpO$_2$, with sleep stages and respiratory events manually scored.
SHHS is widely used for sleep staging and sleep-disordered breathing research, and access requires approval through the National Sleep Research Resource (NSRR).

\noindent
\paragraph{Multi-Ethnic Study of Atherosclerosis (MESA)~\cite{mesa}.}
MESA is a longitudinal NHLBI cohort originally consisting of 6{,}814 Black, White, Hispanic, and Chinese-American adults aged 45--84.
Between 2010 and 2012, 2{,}237 participants completed the MESA Sleep Exam, which included unattended overnight PSG, 7-day wrist actigraphy, and sleep questionnaires.
The dataset provides PSG recordings with rich multi-ethnic demographic and cardiovascular phenotyping, and is available through approved data access requests.

%%%%%%%%%%%%%%%%%%%%%%%%%
\subsection{Architecture and Pre-training}

Electrocardiogram (ECG) and thoracic respiratory effort signals are extracted from each dataset and converted into cardiac waveforms (CW) and breathing waveforms (BW), respectively.
After filtering, normalization, and 30-second patchification, the waveforms are encoded through convolutional layers and a Transformer encoder to produce per-epoch representations.
The model is pre-trained on 240-epoch, 2-hour windows using four-class cross-entropy loss with the AdamW optimizer.

%%%%%%%%%%%%%%%%%%%%%%%%%
\subsection{Transfer to KVSS}

The KVSS dataset provides NIR video recordings with time-aligned contact-sensor signals.
SleepVST requires reliable cardiac and respiratory waveforms, but direct recovery of these waveforms from KVSS videos is challenging.
Facial regions are blurred for privacy, which limits recovery of subtle facial color changes needed for rPPG-based cardiac waveform estimation.
We also attempted to extract respiratory waveforms from NIR video using two representative video-based respiratory estimation methods, phase-based motion magnification~\cite{wadhwa2013phase} and a camera-based respiratory extraction method~\cite{wang2022algorithmic}.
On KVSS, the reconstructed waveforms showed substantial deviations in dominant frequency and amplitude compared with contact thoracic-belt signals, especially under bedding, position changes, weak motion, and non-respiratory body movement.
Consequently, a faithful video-only implementation of the original SleepVST pipeline could not be realized under KVSS conditions.

To enable a controlled comparison on KVSS, we evaluated two oracle-input variants of SleepVST.
In the partial-oracle variant, the cardiac waveform was replaced by the corresponding contact-sensor signal from PSG, while the breathing waveform was video-extracted using phase-based motion magnification~\cite{wadhwa2013phase}.
In the full-oracle variant, both cardiac and breathing waveforms were obtained directly from contact sensors.
These variants should be interpreted as oracle-input versions of one representative proxy-first architecture, not as universal upper bounds for physiological-signal-based sleep staging or practical video-only baselines.

For both variants, the preprocessed waveforms are passed through the frozen SleepVST encoder and Transformer backbone to obtain per-epoch signal features.
To incorporate motion information from video, optical flow is estimated using the DIS (Dense Inverse Search) algorithm~\cite{kroeger2016fast}, and the resulting motion feature vector is concatenated with the SleepVST signal feature.
A Random Forest classifier is then trained on the combined features for four-class sleep staging on KVSS.

%%%%%%%%%%%%%%%%%%%%%%%%%
%%%%%%%%%%%%%%%%%%%%%%%%%
\section{Additional Experiments and Analyses}
\label{appendix:add_res}

%%%%%%%%%%%%%%%%%%%%%%%%%
\subsection{\proposal{} Results}
\label{appendix:overall_res}

\begin{table}[ht]
\centering
\caption{
Detailed class-wise performance of \proposal{}.
(a) Five-class evaluation.
(b) Four-class evaluation after merging N1/N2 into Light and treating N3 as Deep.
}
\vspace{0.5ex}
\label{tab:overall_results}
\small
\renewcommand{\arraystretch}{0.82}

\makebox[0.96\linewidth][c]{%
\begin{minipage}[t]{0.46\linewidth}
\centering
\textbf{(a) 5-Class Classification}\par\vspace{0.5ex}
\begin{tabular*}{\linewidth}{@{\extracolsep{\fill}}lccc@{}}
\toprule
Class & Prec. & Rec. & F1 \\
\midrule
Wake & 0.83 & 0.73 & 0.78 \\
N1   & 0.58 & 0.57 & 0.58 \\
N2   & 0.69 & 0.75 & 0.72 \\
N3   & 0.67 & 0.68 & 0.67 \\
REM  & 0.82 & 0.83 & 0.82 \\
\midrule
Macro Avg.    & 0.72 & 0.71 & \textbf{0.72} \\
Weighted Avg. & 0.72 & 0.71 & 0.72 \\
\midrule
Acc. & \multicolumn{3}{c}{\textbf{0.71}} \\
$\kappa$ & \multicolumn{3}{c}{\textbf{0.62}} \\
\bottomrule
\end{tabular*}
\end{minipage}
\hspace{0.04\linewidth}
\begin{minipage}[t]{0.46\linewidth}
\centering
\textbf{(b) 4-Class Classification}\par\vspace{0.5ex}
\begin{tabular*}{\linewidth}{@{\extracolsep{\fill}}lccc@{}}
\toprule
Class & Prec. & Rec. & F1 \\
\midrule
Wake  & 0.83 & 0.73 & 0.78 \\
Light & 0.82 & 0.86 & 0.83 \\
\phantom{N2} & \phantom{0.00} & \phantom{0.00} & \phantom{0.00} \\
Deep  & 0.67 & 0.68 & 0.67 \\
REM   & 0.82 & 0.83 & 0.82 \\
\midrule
Macro Avg.    & 0.78 & 0.77 & \textbf{0.78} \\
Weighted Avg. & 0.81 & 0.80 & 0.80 \\
\midrule
Acc. & \multicolumn{3}{c}{\textbf{0.80}} \\
$\kappa$ & \multicolumn{3}{c}{\textbf{0.68}} \\
\bottomrule
\end{tabular*}
\end{minipage}
}

\vspace{0.3ex}
{\scriptsize\itshape
In the four-class setting, Light denotes N1+N2 and Deep denotes N3.
}
\end{table}

Table~\ref{tab:overall_results} reports the detailed class-wise performance of \proposal{}.
The model is trained with five AASM sleep-stage labels (Wake, N1, N2, N3, REM).
For the four-class evaluation used in the main paper, both predictions and ground-truth labels are mapped by merging N1 and N2 into Light sleep and treating N3 as Deep sleep.
Thus, the 5-class and 4-class results are obtained from the same trained model, with only the evaluation label mapping changed.

%%%%%%%%%%%%%%%%%%%%%%%%%
\subsection{Backbone Selection for Subject-Relative Micro-Motion Representation}
\label{appendix:backbone_ablation}

\begin{table}[ht]
\centering
\caption{
Backbone selection for the subject-relative micro-motion representation module.
All models are trained with the same pre-spatial micro-motion encoding and within-subject stage contrast.
}
\vspace{0.5ex}
\label{tab:arch_comparison}
\setlength{\tabcolsep}{10pt}
\renewcommand{\arraystretch}{1} 
\small
\begin{tabular}{lccc} 
\toprule
\textbf{Model} & \textbf{Param.} & \textbf{Acc.} & \textbf{MF1} \\
\midrule
Video Swin-Tiny~\cite{swin}     & 28.3M & 0.51 & 0.46 \\
MoViNet-A5~\cite{kondratyuk2021movinets}    & 17.4M & 0.60 & 0.53 \\ 
MoViNet-A2~\cite{kondratyuk2021movinets}    & 4.0M  & \textbf{0.60} & \textbf{0.56} \\
\bottomrule
\end{tabular}
\end{table}

To select the video backbone for the subject-relative micro-motion representation module, we evaluated three candidate architectures: Video Swin-Tiny~\cite{swin}, MoViNet-A5, and MoViNet-A2~\cite{kondratyuk2021movinets}.
All models in this comparison are trained with the same pre-spatial micro-motion encoding and within-subject stage contrast, so the results isolate the effect of the backbone architecture within the representation module.
Table~\ref{tab:arch_comparison} summarizes model size and classification performance.

Video Swin-Tiny performs worse than the MoViNet variants, suggesting that generic attention-based video representations do not directly solve NIR sleep staging.
This may reflect the mismatch between action-oriented video representations and NIR sleep video, where discriminative evidence is weak, localized, and easily dominated by static subject or bedding cues.
The convolutional inductive bias of MoViNet provides a more suitable local spatiotemporal prior for this setting.

Among the MoViNet variants, MoViNet-A2 provides the best accuracy-efficiency tradeoff. 
Although MoViNet-A5 has more than four times the parameters of MoViNet-A2, it does not improve accuracy and yields lower MF1. 
This suggests that increasing generic backbone capacity alone is insufficient for NIR sleep video, where the key challenge is preserving weak, localized motion evidence under strong subject- and acquisition-specific variability. 
We therefore select MoViNet-A2 as the backbone of \proposal{}, balancing performance, model size, and practical deployment constraints.

%%%%%%%%%%%%%%%%%%%%%%%%%
\subsection{Stratified Performance across Recording Conditions and Subject Characteristics}
\label{appendix:stratified_perf}

\begin{figure}[ht]
    \centering
    \includegraphics[width=0.8\textwidth]{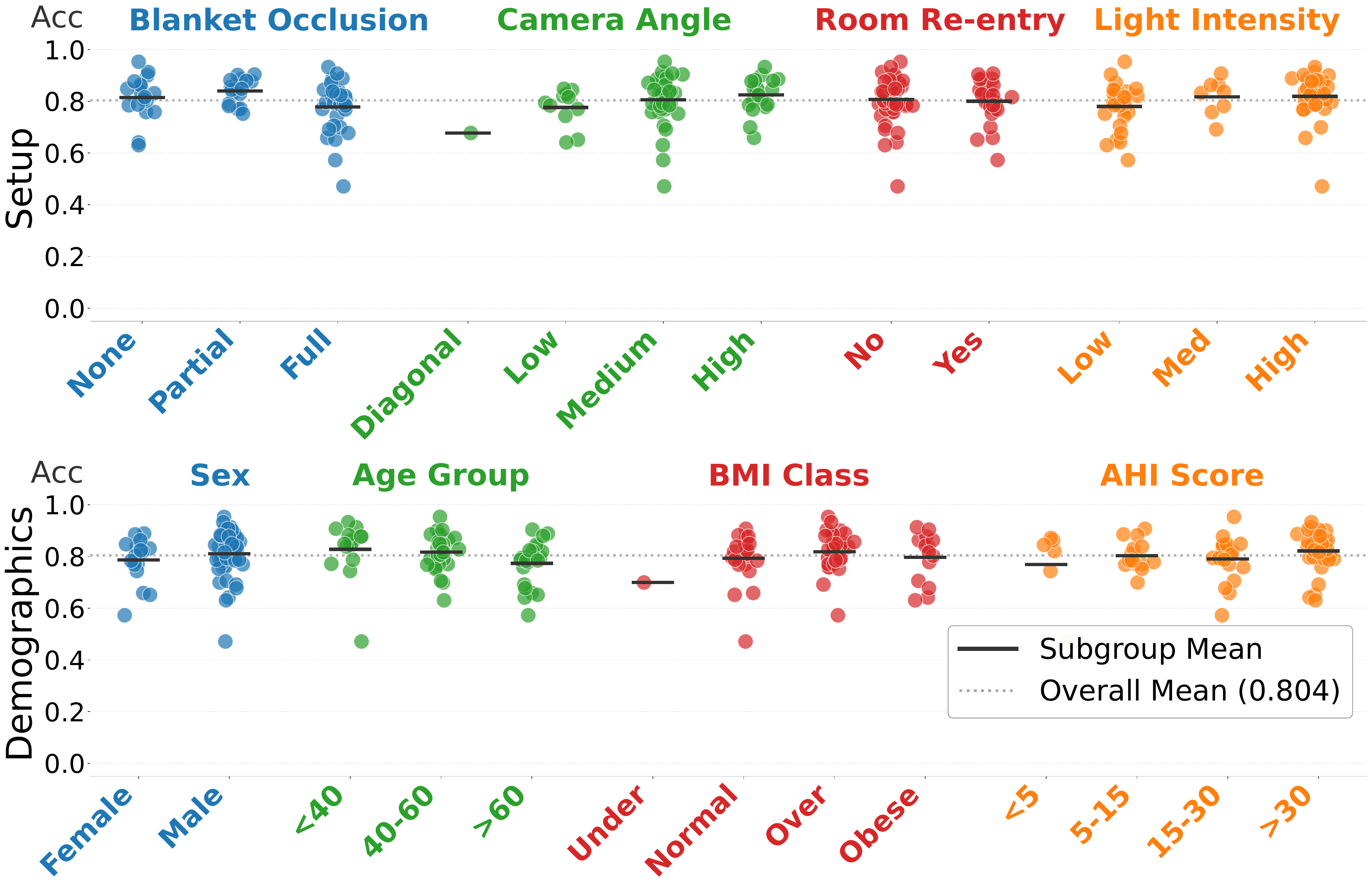}
    \caption{
    Per-subject accuracy stratified by recording conditions and subject characteristics on the held-out test set.
    Each dot represents one subject.
    The upper row shows recording-condition groups, including blanket occlusion, camera angle, room re-entry, and light intensity.
    The lower row shows subject-characteristic groups, including sex, age, BMI class, and AHI group.
    }
    \label{fig:supple_robustness}
\end{figure}

To assess performance consistency under deployment-relevant variation, we conduct a stratified analysis on the held-out test set.
Subjects are grouped by recording condition, including blanket occlusion, camera angle, room re-entry, and light intensity, and by subject characteristic, including sex, age group, BMI class, and AHI group.
Figure~\ref{fig:supple_robustness} reports per-subject accuracy for each subgroup.
\proposal{} maintains comparable performance across the observed subgroups without clear systematic degradation, suggesting stable behavior across the heterogeneous recording environments and patient characteristics represented in KVSS.

%%%%%%%%%%%%%%%%%%%%%%%%%
\subsection{Additional Experiments on Full-Night Sleep Dynamics Modeling}
\label{appendix:ablation_sdm}

%%%%%%%%%%%
\subsubsection{Comparison with alternative sequence models.}

We compare the full-night sequence modeling component of \proposal{} with temporal backbones from four families: recurrent neural networks (RNNs), Transformer-based models, convolutional neural networks (CNNs), and state-space models (SSMs).
In this experiment, each 30-second epoch is treated as one timestamp in the full-night sequence.
All inter-epoch-only sequence models receive the same epoch-level representations, so their comparison isolates the effect of the inter-epoch temporal model.
The ``Mamba (Long only)'' row corresponds to the bidirectional LongMamba path without the within-epoch ShortMamba summarizer, whereas ``SDM (Short+Long)'' reports our full two-scale design.

Table~\ref{tab:sequence_model_comparison} reports overall and per-class results. 
Mamba (Long only) achieves the strongest performance among inter-epoch-only sequence models, with 0.78 accuracy, 0.74 MF1, and 0.65 $\kappa$.
BiLSTM is the closest alternative, while vanilla Transformer, PatchTST, and ModernTCN perform worse under the same epoch-level input setting.
This suggests that full-night sleep staging benefits from sequence models that maintain and update compact temporal states over the overnight trajectory, and supports Mamba as an efficient state-space backbone for long-sequence modeling.

Per-class results show that Mamba (Long only) outperforms the other inter-epoch-only sequence models across all classes, with the clearest advantage on Deep sleep.
The full SDM further improves every class over LongMamba alone, especially Deep sleep, where F1 increases from 0.59 to 0.67.
These results support our design choice: LongMamba is effective for inter-epoch full-night modeling, while ShortMamba improves the epoch-level motion summaries passed to the full-night model.

% \begin{table}[ht]
% \centering
% \caption{
% Overall comparison of sequence models for full-night sleep dynamics modeling.
% All models operate on epoch-level representations, where each 30-second epoch is treated as one timestamp.
% Best results are shown in \textbf{bold}, and second-best results are shown in \textit{italics}.
% }
% \vspace{1ex}
% \label{tab:overall_sequence_model_comparison}
% \begin{tabular}{l l c c c}
% \toprule
% \textbf{Model} & \textbf{Family} & \textbf{Acc.} & \textbf{MF1} & \textbf{\(\kappa\)} \\
% \midrule
% BiLSTM~\cite{huang2015bidirectional} & RNN & 0.77 & 0.72 & 0.62 \\
% PatchTST~\cite{Yuqietal-2023-PatchTST} & Transformer & 0.74 & 0.70 & 0.58 \\
% ModernTCN~\cite{donghao2024moderntcn} & CNN & 0.73 & 0.68 & 0.56 \\
% Transformer Encoder~\cite{vaswani2017attention} & Transformer & 0.71 & 0.62 & 0.52 \\
% Mamba (Long only)~\cite{gu2023mamba} & SSM & \textit{0.78} & \textit{0.74} & \textit{0.65} \\
% \textbf{SDM (Short+Long)}~\cite{gu2023mamba} & SSM & \textbf{0.80} & \textbf{0.78} & \textbf{0.68} \\
% \bottomrule
% \end{tabular}

% {\scriptsize\itshape
% RNN: recurrent neural network; CNN: convolutional temporal model; SSM: state-space model. \\
% All models operate on epoch-level full-night sequences.
% \par}

% \end{table}

\begin{table*}[t]
\centering
\caption{
Comparison of sequence models for full-night sleep dynamics modeling.
All models operate on epoch-level representations, where each 30-second epoch is treated as one timestamp.
Best results are shown in \textbf{bold}, and second-best results are shown in \textit{italics}.
}
\label{tab:sequence_model_comparison}
\vspace{0.5ex}
\small
\setlength{\tabcolsep}{4.2pt}
\renewcommand{\arraystretch}{0.9}
\begin{tabular}{l l c c c c c c c}
\toprule
\multirow{2}{*}{\textbf{Model}} &
\multirow{2}{*}{\textbf{Family}} &
\multicolumn{3}{c}{\textbf{Overall}} &
\multicolumn{4}{c}{\textbf{Per-class F1}} \\
\cmidrule(lr){3-5} \cmidrule(lr){6-9}
& & \textbf{Acc.} & \textbf{MF1} & \(\boldsymbol{\kappa}\) 
& \textbf{Wake} & \textbf{Light} & \textbf{Deep} & \textbf{REM} \\
\midrule
BiLSTM~\cite{graves2005framewise} 
& RNN 
& 0.77 & 0.72 & 0.62 
& 0.76 & 0.80 & 0.53 & \textit{0.80} \\

PatchTST~\cite{Yuqietal-2023-PatchTST} 
& Transformer 
& 0.74 & 0.70 & 0.58 
& 0.71 & 0.78 & 0.53 & 0.76 \\

ModernTCN~\cite{donghao2024moderntcn} 
& CNN 
& 0.73 & 0.68 & 0.56 
& 0.70 & 0.77 & 0.50 & 0.75 \\

Vanilla Transformer~\cite{vaswani2017attention} 
& Transformer 
& 0.71 & 0.62 & 0.52 
& 0.72 & 0.77 & 0.33 & 0.66 \\

Mamba (Long only)~\cite{gu2023mamba} 
& SSM 
& \textit{0.78} & \textit{0.74} & \textit{0.65} 
& \textit{0.77} & \textit{0.81} & \textit{0.59} & \textit{0.80} \\

\textbf{SDM (Short+Long)}~\cite{gu2023mamba} 
& SSM 
& \textbf{0.80} & \textbf{0.78} & \textbf{0.68} 
& \textbf{0.78} & \textbf{0.83} & \textbf{0.67} & \textbf{0.82} \\
\bottomrule
\end{tabular}

\vspace{0.3ex}
{\scriptsize\itshape
RNN: recurrent neural network; CNN: convolutional temporal model; SSM: state-space model.
Light denotes N1+N2 and Deep denotes N3.
\par}
\vspace{-0.5ex}
\end{table*}

\noindent
\textbf{Implementation details of compared sequence models.}
All compared models are trained to produce per-epoch five-class logits (Wake, N1, N2, N3, REM); reported four-class results are obtained by merging N1 and N2 into Light and treating N3 as Deep.

\textit{BiLSTM.}
We use a two-layer bidirectional LSTM~\cite{graves2005framewise}  with hidden size 256.
Epoch representations are projected to 256 dimensions using LayerNorm and a linear layer, processed over the full-night sequence, and decoded by a two-layer MLP head at each epoch.

\textit{Vanilla Transformer.}
We use a vanilla Transformer~\cite{vaswani2017attention}  with three encoder layers, \(d_{\mathrm{model}}=256\), 8 attention heads, \(d_{\mathrm{ff}}=1024\), pre-normalization, GELU activations, and dropout 0.1.
This baseline models the full-night epoch sequence with bidirectional self-attention and a per-epoch MLP classification head.

\textit{PatchTST.}
We adapt PatchTST~\cite{Yuqietal-2023-PatchTST} to epoch-level sequence classification.
Patches are formed over consecutive epoch-level timestamps, not raw video frames, using \(\text{patch\_len}=16\) and \(\text{stride}=8\).
We use the channel-independent TSTiEncoder with \(d_{\mathrm{model}}=128\), 16 heads, 3 layers, and \(d_{\mathrm{ff}}=256\).
Because PatchTST is originally designed for forecasting, we replace the forecasting head with a classification head that maps patch-level outputs back to epoch-level predictions by overlap averaging.

\textit{ModernTCN.}
We use a ModernTCN~\cite{donghao2024moderntcn} encoder with ETTh1-style hyperparameters and attach a per-epoch classification head.
The encoder uses RevIN, a stem layer, and one ModernTCN stage with large-kernel depthwise temporal convolution and grouped ConvFFN blocks.
The output is compressed and decoded by a two-layer MLP head at each epoch.

\textit{Mamba (Long only).}
The Long-only Mamba baseline uses the full-night LongMamba path without the within-epoch ShortMamba summarizer.
It consists of four bidirectional LongMamba blocks with \(d_{\mathrm{model}}=128\) and \(d_{\mathrm{state}}=32\), using epoch-index embeddings and a shared per-epoch classifier.

\textit{SDM (Short+Long).}
The full SDM uses both ShortMamba and LongMamba.
ShortMamba first summarizes within-epoch motion evolution into compact epoch tokens, and LongMamba then models the stacked epoch tokens across the full night.

%%%%%%%%%%%
\subsubsection{Effect of temporal context length.}

To examine how temporal context length affects full-night sleep dynamics modeling, we train the Mamba (Long only) backbone with context windows of 
\(W \in \{10, 30, 50, 100, 200, 500, \text{full}\}\), while keeping all other hyperparameters fixed.
Each 30-second epoch is treated as one timestamp, and each training iteration samples \(K=5\) random length-\(W\) crops per subject.
When a recording contains fewer than \(W\) epochs, zero padding is used with \texttt{ignore\_index} masking.
At inference, we use overlapping sliding windows with stride \(W/10\) and aggregate predictions by averaging logits for epochs covered by multiple windows.
This setup isolates the effect of temporal context length while keeping the sequence model and training recipe fixed.

\begin{table}[ht]
\centering
\caption{
Effect of temporal context length on full-night sleep dynamics modeling.
Each 30-second epoch is treated as one timestamp.
}
\label{tab:window_size_ablation}
\begin{tabular}{c c c c}
\toprule
\textbf{Window size} & \textbf{Acc.} & \textbf{MF1} & \textbf{\(\kappa\)} \\
\midrule
10 & 0.72 & 0.64 & 0.53 \\
30 & 0.74 & 0.67 & 0.58 \\
50 & 0.76 & 0.71 & 0.61 \\
100 & 0.77 & 0.72 & 0.63 \\
200 & 0.77 & 0.73 & 0.63 \\
500 & 0.78 & 0.74 & 0.64 \\
\textbf{full} & \textbf{0.78} & \textbf{0.74} & \textbf{0.65} \\
\bottomrule
\end{tabular}
\end{table}

Table~\ref{tab:window_size_ablation} shows that longer context consistently improves sleep dynamics modeling.
Accuracy increases from 0.72 at \(W=10\) to 0.78 with full-night context, while MF1 improves from 0.64 to 0.74 and \(\kappa\) from 0.53 to 0.65.
The largest gains occur as the window grows up to \(W=100\), after which accuracy begins to plateau.
However, MF1 and \(\kappa\) continue to improve with longer context, suggesting that full-night information is especially helpful for less frequent or temporally structured stages such as Deep and REM.

%%%%%%%%%%%%%%%%%%%%%%%%%
\subsection{Example Full-Night Test Predictions}
\label{appendix:example_predictions}

\begin{figure}[ht]
    \centering
    \includegraphics[width=1.0\textwidth]{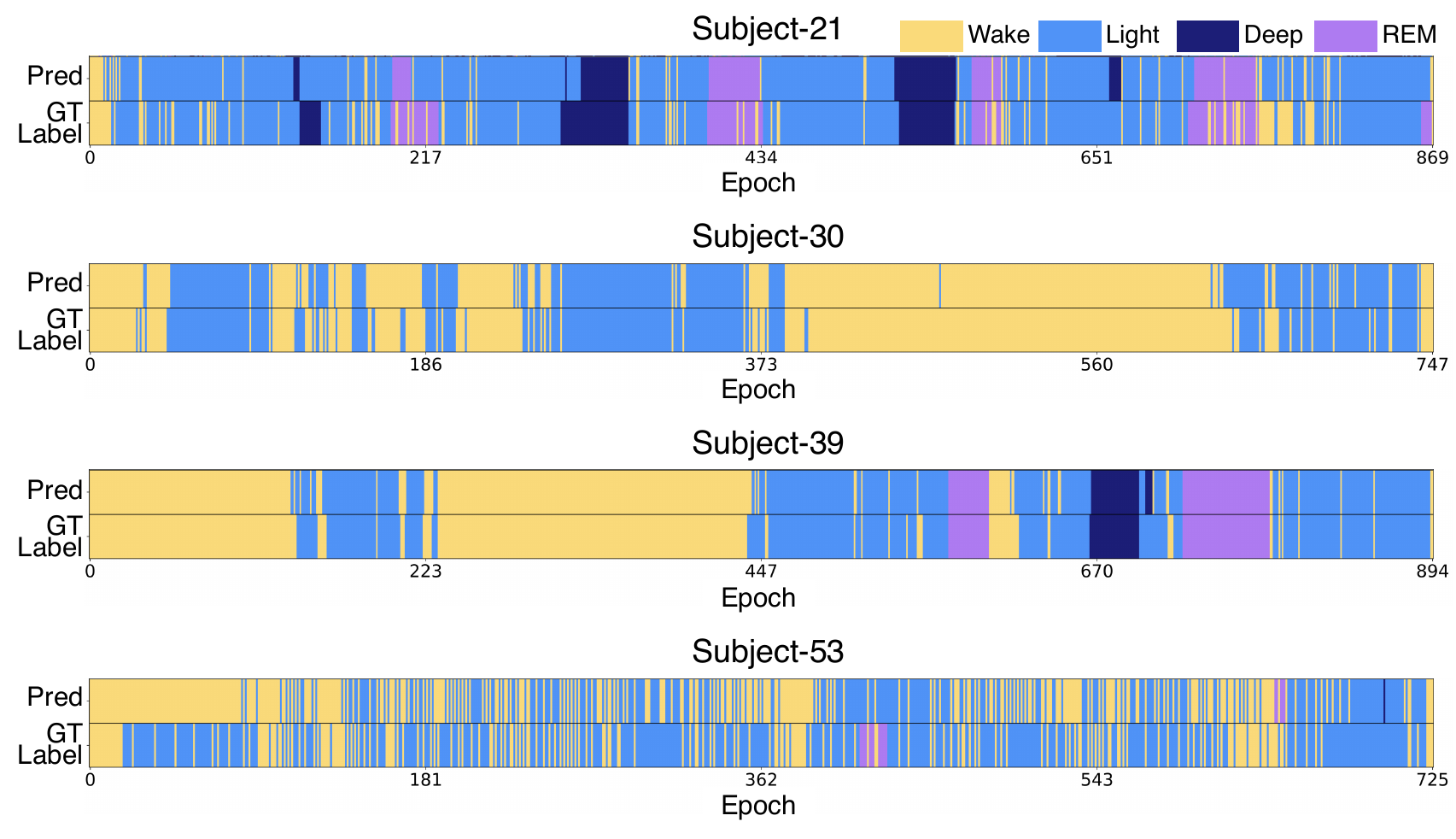}
    \caption{
    Example full-night predictions on heterogeneous test subjects.
    Ground-truth (GT) labels and model predictions are shown for four representative subjects with different sleep-stage trajectories.
    }
    \label{fig:supple_test_example}
\end{figure}

Figure~\ref{fig:supple_test_example} provides additional full-night prediction examples complementing Figure~\ref{fig:full-night}.
The selected subjects exhibit heterogeneous sleep architectures, ranging from multi-cycle sleep patterns to irregular profiles dominated by Wake and Light sleep.
Subject~21 shows multiple stage transitions across the night, and the model follows the changing proportions of Light, Deep, and REM stages without collapsing to a fixed template.
Subjects~30, 39, and 53 illustrate more atypical patterns, including recordings with little or no Deep/REM sleep, prolonged wakefulness, and frequent Wake--Light alternations.
Across these examples, most prediction errors occur near stage boundaries or short transient segments, while longer sleep-stage structures are largely preserved.
These cases further illustrate that full-night modeling helps \proposal{} track subject-specific overnight trajectories under diverse sleep patterns.

%%%%%%%%%%%%%%%%%%%%%%%%%
%%%%%%%%%%%%%%%%%%%%%%%%%
\section{Sleep Video EigenCAM (SVECAM)}
\label{appendix:svecam}

\begin{figure*}[ht!]
    \centering
    \includegraphics[width=.97\linewidth]{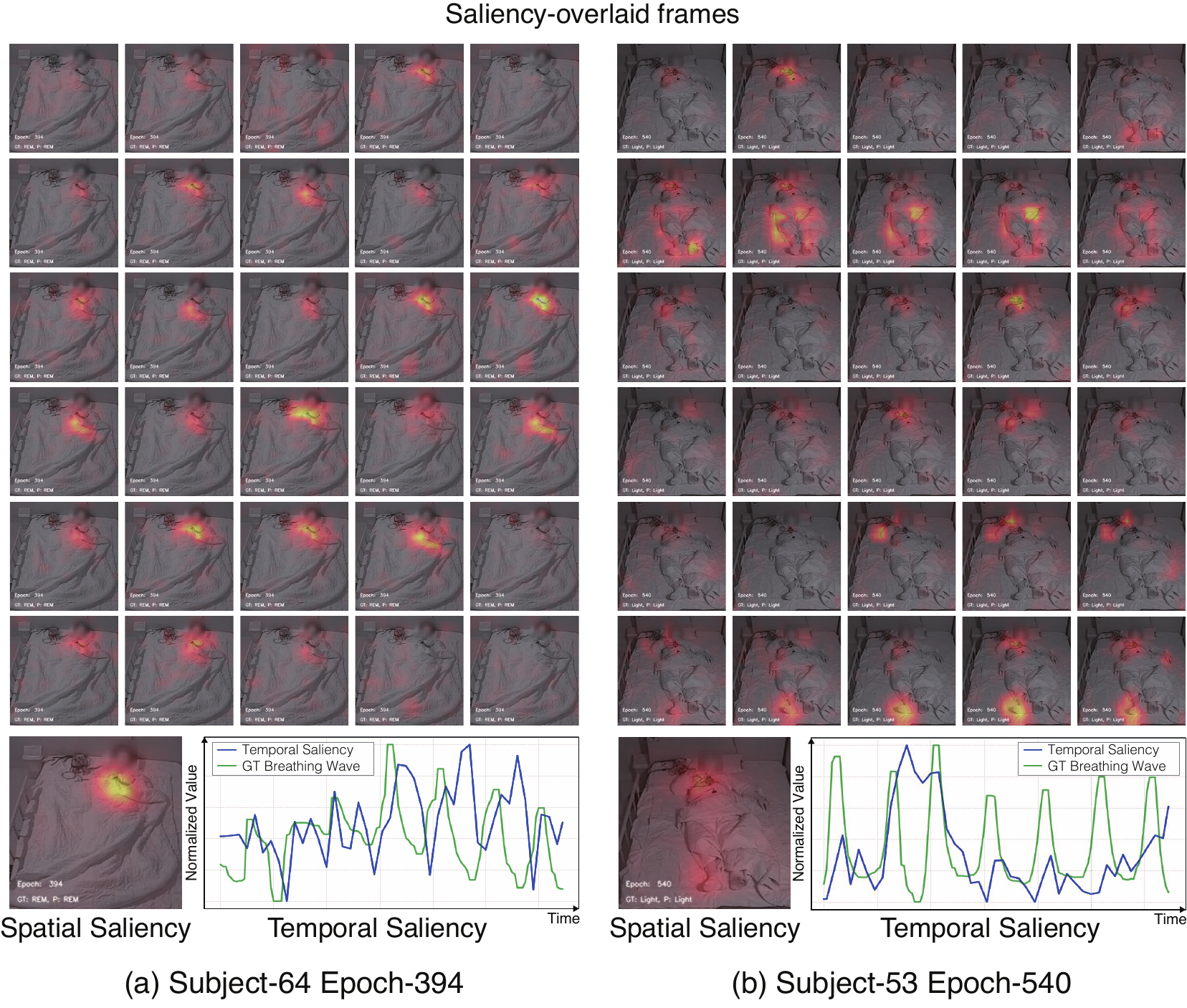}
    \caption{
    SVECAM visualizations for two representative NIR sleep-video epochs.
    For each epoch, we show spatial saliency maps, temporal saliency curves, and saliency-overlaid frames.
    Frames are sampled at 1\,Hz over the 30\,s epoch and arranged sequentially from left to right, wrapping every five seconds.
    }
    \label{fig:supple_eigencam}
\end{figure*}

NIR sleep videos provide weak and indirect visual evidence for sleep-stage discrimination, making it important to inspect which spatiotemporal regions the model uses.
We therefore extend EigenCAM~\cite{eigencam} to 3D activation tensors from sleep-video models.

Given an activation tensor 
\(\mathbf{A}\in\mathbb{R}^{C\times T\times H\times W}\), 
a direct EigenCAM application would flatten temporal and spatial dimensions before singular value decomposition (SVD), which can mix motion-related and appearance-related structure.
We instead use Sleep Video EigenCAM (SVECAM), which derives temporal and spatial channel weights separately and then combines them.

\begin{itemize}[leftmargin=*, noitemsep, topsep=0pt]
    \item \textbf{Temporal weighting.}
    For each spatial location \((h,w)\), we apply SVD to 
    \(\mathbf{A}_{h,w}\in\mathbb{R}^{C\times T}\) 
    and use the leading left singular vector 
    \(\mathbf{u}_T^{(h,w)}\in\mathbb{R}^{C}\) 
    as a channel-weight vector capturing temporal co-variation at that location.

    \item \textbf{Spatial weighting.}
    For each frame \(t\), we apply SVD to 
    \(\mathbf{A}_t\in\mathbb{R}^{C\times(H\cdot W)}\) 
    and use the leading left singular vector 
    \(\mathbf{u}_S^{(t)}\in\mathbb{R}^{C}\) 
    as a channel-weight vector capturing spatially coherent activation at that time.
\end{itemize}

We form a spatiotemporal weight tensor 
\(\mathbf{W}\in\mathbb{R}^{C\times T\times H\times W}\) 
by broadcasting and summing the normalized temporal and spatial channel weights:
\[
\mathbf{W}(c,t,h,w)
= \tilde{\mathbf{u}}_T^{(h,w)}(c)
+ \tilde{\mathbf{u}}_S^{(t)}(c),
\]
where \(\tilde{\mathbf{u}}\) denotes an \(\ell_2\)-normalized vector.
The final SVECAM saliency is computed as a channel-weighted activation map followed by ReLU:
\[
\mathrm{SVECAM}(t,h,w)
=
\mathrm{ReLU}
\left(
\sum_c
\mathbf{A}_{c,t,h,w}
\mathbf{W}_{c,t,h,w}
\right).
\]

This formulation highlights activation channels that are temporally informative at each spatial location and spatially coherent within each frame.
Because NIR sleep videos are captured from a fixed camera and the body layout is relatively stable within many 30-second epochs, preserving this spatiotemporal structure is useful for interpreting localized sleep-related motion.

SVECAM is applied to the subject-relative micro-motion representation module.
Figure~\ref{fig:supple_eigencam} presents visualizations for representative 30\,s epochs from different subjects.
In example~(a), saliency is concentrated around the thoraco-abdominal region, and the temporal saliency curve follows respiration-related periodic motion.
This suggests that the model uses breathing-related visual motion without explicitly reconstructing a respiratory waveform.

In example~(b), the epoch contains transient body movements in addition to respiration-related motion.
SVECAM highlights larger movements when they occur, while saliency returns to the thoraco-abdominal region during quieter intervals.
These examples suggest that \proposal{} uses both localized respiration-related motion and transient behavioral motion as visual evidence for sleep staging.

\end{document}